\documentclass{article}
\usepackage{arxiv}

\usepackage[utf8]{inputenc} 
\usepackage[T1]{fontenc}    
\usepackage{tabularx}
\usepackage{array}
\usepackage{float}
\usepackage{algorithm}
\usepackage{algorithmic}
\usepackage{hyperref}       
\usepackage{url}            
\usepackage{booktabs}       
\usepackage{amsfonts}       
\usepackage{amsmath}        
\usepackage{amssymb}        
\usepackage{subfigure}      
\usepackage{nicefrac}       
\usepackage{multirow}       
\usepackage{microtype}      
\usepackage{lipsum}		
\usepackage{graphicx}
\usepackage{doi}
\usepackage{float}
\DeclareMathOperator{\softmax}{softmax}

\usepackage{natbib}
 \bibpunct[, ]{(}{)}{,}{a}{}{,}%
\makeatletter

\renewcommand{\normalsize}{\@setfontsize\normalsize{11}{13}}

\renewcommand{\small}{\@setfontsize\small{10.3}{12}}
\renewcommand{\footnotesize}{\@setfontsize\footnotesize{9.7}{11}}
\makeatother

\title{Graph Optimization Foundation Model: Tokenizing Graph via A Language-Model Paradigm}

\author{
  Yunhao Liang$^{1}$ \quad
  Pujun Zhang$^{1}$\thanks{Correspondence to: Pujun Zhang (pjzhang@hku.hk), Yuan Qu (yuanqu@hku.hk)} \quad
  Yuan Qu$^{1}$\footnotemark[1] \quad
  Jingyuan Yang$^{2}$ \quad
  Shaochong Lin$^{1}$ \quad
  Max Z.J. Shen$^{1,3,4}$ \\
  \\
  $^{1}$Faculty of Engineering, The University of Hong Kong, Hong Kong SAR, China \\
  $^{2}$Costello College of Business, George Mason University, VA, USA \\
  $^{3}$Faculty of Business and Economics, The University of Hong Kong, Hong Kong, China \\
  $^{4}$College of Engineering, University of California, Berkeley, CA, USA
}

\renewcommand{\shorttitle}{\textit{arXiv} Template}

\hypersetup{
pdftitle={A template for the arxiv style},
pdfsubject={q-bio.NC, q-bio.QM},
pdfauthor={David S.~Hippocampus, Elias D.~Striatum},
pdfkeywords={First keyword, Second keyword, More},
}
\date{}

\begin{document}
\maketitle
\vspace{-2em}






\begin{abstract}
    Graph optimization is foundational to modern operations, routing vehicles through road networks, detecting communities in social networks, and so on.
Yet, prevailing paradigms, ranging from classical heuristics to recent neural solvers, treat each task instance in isolation. 
This solve-from-scratch paradigm overlooks the persistent topological knowledge embedded within the underlying network, leading to significant knowledge loss across recurring decision cycles and multiple optimization problems, which usually results in computational redundancy, limits in scalability, and real-world infeasibility.
Therefore, we introduce the Graph Optimization Foundation Model (GOFM), which shifts the focus from task-specific solvers to the pretraining of a task-agnostic structural prior. GOFM internalizes network topology and distance geometry by encoding structure-aware random walks into a Transformer through progressive masked reconstruction. This process tokenizes a representation that captures the fundamental properties of the graph. At inference, this learned prior facilitates diverse optimization tasks via lightweight constrained decoding, ensuring strict feasibility without retraining. Evaluations across five routing families and networks of up to 1,945 nodes demonstrate that GOFM achieves competitive performance compared to specialized solvers while significantly reducing computational time. Beyond routing, a real-world Amazon Last Mile case study and experiments on community detection and influence maximization confirm the model's capacity for representation reuse. 
For practitioners, GOFM offers a path toward treating persistent networks as reusable decision assets, reducing repeated solver engineering while enabling fast, feasible decision support at scale.
\end{abstract}



\maketitle

%

\section{Introduction}
In Greek mythology, Sisyphus is condemned to roll a boulder up a hill for eternity. Each time he nears the summit, the boulder rolls back, and he must begin again from nothing. The punishment is not the effort itself but the absence of memory: no ascent teaches him anything that makes the next one easier. Network optimization today suffers from a strikingly similar curse. Every time a new routing, partitioning, or targeting problem arrives on the same graph, the solver starts from the bottom of the hill again and again. 
The connection patterns it reconstructed yesterday, the bottlenecks it discovered last week, and the distance relationships it computed an hour ago, all the knowledge is not carried forward while the graph remains the same. 
If this persistent structural knowledge could be distilled from the graph, must every downstream optimization task still begin from the bottom of the hill?

The answer may lie in the pretrain-transfer paradigm that has recently redefined artificial intelligence~\citep{Bommasani2021OnTO, Brown2020LanguageMA}. 
Large language models (LLMs) learn structured regularities from large-scale data through self-supervision.
They then transfer that knowledge across tasks with minimal adaptation~\citep{BERT, Achiam2023GPT4TR, Yang2025Qwen3TR}. 
The model sequentially learns which completions are plausible in a given context, and the resulting prior, a form of learned background knowledge, transfers to tasks the model was never explicitly trained for. 

A similar ``recipe'' is both possible and necessary for graph optimization.
Graph optimization problems are central to Operations Research (OR), spanning routing and logistics on road networks, community structure in social networks, and influence-based targeting in information networks~\citep{Laporte2009FiftyYO,Girvan2001CommunitySI,Jaouadi2024ASO}. 
Many of these problems are NP-hard, meaning that no efficient exact algorithm is known for the worst case, and the classical trade-off is well known: exact methods deliver optimal solutions at prohibitive computational cost, while heuristics achieve tractability with limited optimality guarantees and often require substantial problem-specific engineering \citep{Papadimitriou1981CombinatorialOA}. In real-world industrial settings where the underlying infrastructure graph is persistent and decision queries change frequently, this trade-off is especially costly. Each new task instance triggers a fresh solve that discards all previously acquired structural knowledge. What is missing is a reusable computational artifact that internalizes structural constraints and thereby amortizes the computational cost of learning over the entire lifecycle of the infrastructure.

Realizing this vision, however, requires bridging a significant gap in OR.
A range of neural architectures designed for combinatorial problems, from pointer networks \citep{Vinyals2015PointerN} and reinforcement-learning-trained attention models \citep{Bello2016NeuralCO, Kool2018AttentionLT, Kwon2020POMOPO} to diffusion-based solvers \citep{Sun2023DIFUSCOGD}, typically rely on specialized output modules tailored to one problem type and report strong performance on synthetic, small-scale, or complete-graph settings where every pair of nodes is directly connected. LLM-centric approaches either cast optimization as text generation or use LLMs as modeling and decomposition agents that make automated calls to external solver software \citep{Yang2023LargeLM, Zhang2025ORLLMAgentAM, Huang2024ORLMAC}. However, most of this work is benchmarked on complete-graph or small-scale instances, and many pipelines remain problem-specific through task-tailored decoders, prompting templates, or solver-dependent tool chains. These design choices lead to weak feasibility on real networks, where connection patterns, distance structure, and hard constraints play a central operational role, and they limit cross-task transfer in settings where one graph supports many different problems over time.

To address this gap, we propose the Graph Optimization Foundation Model (GOFM), a pretraining framework that learns a transferable \emph{structural prior}, a learned summary of the graph's connectivity patterns and distance relationships, over a fixed weighted graph. GOFM is pretrained on a \emph{graph corpus}, a large collection of synthetic trajectories sampled from the graph, to internalize these regularities through self-supervision, without using any problem-specific labels or rewards. With this structural prior and lightweight task specifications, GOFM supports graph optimization through downstream decoding, covering distance-based routing problems as its primary focus and extending to non-routing tasks such as community detection and influence maximization as demonstrations of generalizability. The key idea is to replace repeated per-task solver engineering with a single learned backbone that can be steered toward different objectives by changing only the decoding procedure, the mechanism that converts the model's learned knowledge into a task-specific solution.

Concretely, we transform a raw graph into self-supervised training signals through structure-aware random walks that respect both connectivity and edge weights. This walk-based approach is inspired by Node2Vec \citep{Grover2016node2vec}, which showed that converting graph structure into random-walk sequences enables effective representation learning. The resulting trajectories serve as training data for a progressive masked-reconstruction objective: portions of each trajectory are hidden, and the model learns to predict the missing segments, starting with large gaps and refining to shorter ones. In this way, the model learns how nodes connect, which regions are reachable from one another, and how subpaths compose consistently across the network. The resulting backbone is not tied to any single optimization problem. Rather, it encodes reusable structural knowledge that acts as a learned prior over plausible completions of a partial solution at inference. A decoding procedure, together with constraint projection, steers this prior toward tasks such as shortest path, tour-family routing, and subset selection, while maintaining hard feasibility throughout. No task-specific feedback is used during pretraining, in contrast to many neural OR pipelines that rely on supervised labels or reinforcement-learning rewards tied to a single problem \citep{Bello2016NeuralCO}. This separation of universal structural pretraining from thin task-specific generation is what enables cross-task reuse on real networks.

Empirically, the pretrained GOFM backbone adapts to diverse distance-based graph optimization tasks, including multiple NP-hard problem families, without any architectural modification. Across scales from small synthetic graphs ($N{=}20$) to large real networks ($N{=}1,945$), GOFM achieves competitive solution quality compared to widely used OR solvers such as LKH3~\citep{Helsgaun2017AnEO}, a state-of-the-art tour optimizer, and OR-Tools~\citep{ortools}, Google's optimization toolkit, with inference one to two orders of magnitude faster on city-scale graphs. Unlike neural baselines trained on complete-graph distributions, GOFM maintains near-perfect feasibility across all settings. A real-world Amazon Last Mile case~\citep{Merchn20222021AL} study shows a 24.5\% distance reduction without task-specific tuning. Representation-reuse experiments on community detection and influence maximization confirm that the structural prior generalizes beyond routing.

In summary, this paper makes three contributions:

\textbf{(1) A foundation-model framework for realistic, distance-based graph problems.} We introduce GOFM, a single model with lightweight constraint projection that performs well across multiple graph tasks, from shortest path to NP-hard routing variants such as tour-family problems, on real networks without any architectural changes.

 \textbf{(2) Self-supervised pretraining via structure-aware walks and insertion-based reconstruction.} We design a scalable graph-to-corpus mechanism based on structure-aware random walks and a progressive insertion-based reconstruction objective that equips the model with a holistic structural prior over the target graph.


\textbf{(3) Empirical validation of cross-task reuse and practical relevance on real networks.}
Across multiple road-network scales, routing families, and a real-world Amazon Last Mile case study, we show that a pretrained backbone can support diverse downstream decision problems with lightweight task adaptation. These results validate the extensibility of the pretrain--transfer paradigm across tasks and highlight its practical value for repeated decisions on persistent networks.

The remainder of this paper is organized as follows. Section~\ref{sec:lr} reviews related work across four research streams and positions our contribution relative to each. Section~\ref{sec:preliminary} introduces the formal problem setup and traces the conceptual bridge from language-model pretraining to graph optimization. Section~\ref{sec:gofm} details the three-stage GOFM framework: corpus generation, pretraining, and guided decoding. Section~\ref{sec:exp} presents experimental results on four network scales and five routing task families, including the Amazon Last Mile case study. Section~\ref{sec:dis} discusses generalizability beyond routing, managerial implications, and limitations. Section~\ref{sec:conclusion} concludes.


\section{Related Work} \label{sec:lr}

This study connects to four research streams: (i) learning-based combinatorial optimization, particularly routing and tour construction; (ii) graph representation learning for encoding network structure; (iii) LLM-based optimization, including direct solution generation and LLM-assisted modeling; and (iv) graph foundation models that seek transferable graph-native priors across tasks or domains. We review each stream in turn, clarify its relevance to our setting, highlight what changes when the target setting involves sparse, persistent infrastructure graphs, and identify the gap that motivates our graph-native pretrain--decode framework.

\subsection{Learning-based combinatorial optimization for routing and tours}

Learning-based combinatorial optimization has largely focused on building instance-to-solution mappings: given a problem instance, a model outputs a structured decision such as a permutation or a set of routes. Early neural approaches trace back to energy-based dynamics for optimization~\citep{Hopfield1985NeuralCO}. Modern work relies on attention-based encoder-decoder architectures in which a sequence model constructs tours step by step. Pointer Networks~\citep{Vinyals2015PointerN} showed that attention can represent permutation outputs. Subsequent reinforcement-learning formulations reduced the need for supervised optimal labels by using tour length as a reward signal~\citep{Bello2016NeuralCO}. Attention-based routing models further improved scalability and solution quality through constructive decoding trained on synthetic distributions~\citep{Kool2018AttentionLT}, and multi-start strategies such as POMO exploit multiple initializations to improve robustness~\citep{Kwon2020POMOPO}.

Transformers~\citep{Vaswani2017Attention} have become the dominant architecture for neural routing because they effectively aggregate global context. Representative work includes Transformer encoders trained via reinforcement learning~\citep{Bresson2021TheTN}, designs that reduce attention complexity~\citep{Yang2023TSPformer}, and multi-pointer decoding mechanisms~\citep{Jin2023Pointerformer}. Beyond constructive approaches, generative paradigms such as diffusion have also been applied to combinatorial optimization, with DIFUSCO as a representative example \citep{Sun2023DIFUSCOGD}. Broader frameworks seek reusable neural components across problem families \citep{Luo2023NeuralCO}, and recent work explores structural features or instance-conditioned adaptation modules to improve robustness \citep{Zhao2025ATS, Zhou2024InstanceConditionedAF}. Related ``learning to optimize'' pipelines predict intermediate objects followed by post-processing \citep{Khalil2017LearningCO, Deudon2018LearningHF}.

A recurring pattern across this literature, however, is that most methods are developed and benchmarked on Euclidean instances over complete graphs. In that setting, feasibility is implicit and edge costs are metric by construction. Real road networks are fundamentally different: they are sparse and topologically constrained, with costs determined by the network metric rather than Euclidean distance. Policies trained on complete-graph distributions often need additional assumptions (such as metric closure) and show degraded feasibility when applied to sparse infrastructure graphs. This motivates a shift from training a separate solver per task toward a reusable computational artifact that learns the structure of a persistent graph once \citep{bertsimas2020predictive, bengio2021machine}. The Transformer and generative approaches reviewed above motivate our architectural choice, because attention mechanisms flexibly aggregate global context and capture long-range dependencies. But our key difference is the role of the architecture: rather than learning a task-specific policy that directly emits tours or routes, we use Transformer capacity to learn a reusable structural prior on a fixed sparse graph via self-supervised trajectory reconstruction. This connects naturally to the representation-learning stream reviewed next.

\subsection{Graph representation learning for encoding network structure}
\label{sec:lr-graph-repr}

Whereas the first stream learns task-oriented solvers, the second stream focuses on learning representations that encode graph structure for reuse across downstream analyses. 
This stream of research draws its methodological impetus from word embedding techniques—a foundational paradigm shifted from the field of Natural Language Processing (NLP).
The canonical self-supervised recipe for sequence-based representation learning is Word2Vec, which learns token embeddings by context prediction on large unlabeled corpora \citep{Mikolov2013EfficientEO, Mikolov2013DistributedRO}. Two standard variants exist. Continuous Bag-of-Words (CBOW) predicts a target token from its surrounding context, treating context as an order-agnostic bag of neighboring tokens. Skip-gram does the reverse, using a center token to predict surrounding context tokens within a fixed window. Node2Vec transfers this idea to graphs by converting a graph into a trajectory corpus of random-walk ``sentences'' over node ``tokens'' and learning embeddings from co-occurrence patterns. \citep{Grover2016node2vec}.

Graph neural networks such as GCN \citep{Kipf2016GCN}, GraphSAGE \citep{Hamilton2017GraphSage}, and GAT \citep{Velickovic2018GAT} take a different approach, learning permutation-invariant node representations from adjacency structure and node attributes. These models are natural candidates for network representation and have been widely adopted for node classification, link prediction, and graph classification tasks. The shift from solver training to structural pretraining repositions the model as a representation learner rather than a task-specific policy. The same Transformer family that often appears in routing decoders can instead serve as a graph encoder whose outputs summarize topology and metric geometry in a way that is portable across objectives.

However, standard graph embeddings are typically optimized for node-level or edge-level prediction and clustering objectives. They are not designed to model trajectory compatibility under weighted network metrics, which is what governs distance-based decisions. Moreover, representations built for prediction do not by themselves specify how to enforce hard feasibility constraints when constructing paths, tours, or route sets. This gap motivates graph-native pretraining objectives that treat connectivity and metric geometry as the supervisory signal and produce a reusable structural prior over feasible subpath composition on a fixed weighted graph. Such a prior can then be paired with explicit constraint-aware decoding at task time.

\subsection{LLM-based optimization and LLM-assisted modeling}

While the first two streams focus on neural solvers and graph representations respectively, a third stream leverages large language models as optimization engines or as assistants in the OR pipeline. In direct-solving settings, some work embeds graphs into language-model training or prompting interfaces, such as GraphGPT \citep{Tang2023GraphGPTGI}. Optimization-specific pipelines use LLMs in iterative improvement loops \citep{Yang2023LargeLM} or as evolutionary operators \citep{Liu2023LargeLM}. More specialized approaches include autoregressive solvers trained with preference learning for tour-as-text formulations \citep{Ghimire2025OneShotAG} and multimodal variants that use visual instance representations for small-scale routing \citep{Elhenawy2024VisualRA}.

A complementary line of work positions LLMs as assistants for OR workflows. This includes translating natural-language specifications into optimization models or code \citep{Huang2024ORLMAC, huang2024words} and structured prompting with tool use \citep{Zhang2025ORLLMAgentAM}. Heuristic synthesis with reflective evolution is another emerging direction \citep{Ye2024ReEvoLL}. These approaches expand the role of learning in OR, but they share a common limitation: they do not provide a graph-native, transferable representation that directly encodes sparse topology and weighted geometry. Language-token representations are not naturally aligned with graph feasibility constraints. As a result, performance tends to degrade as instance size and constraint complexity increase. Our framework addresses this representation mismatch by learning a structural prior directly on the graph and coupling it with explicit constraint-aware decoding for downstream optimization tasks.
\subsection{Graph foundation models}

A closely related and rapidly growing stream studies graph foundation models (GFMs), which aim to learn general-purpose graph representations that transfer across datasets, domains, and task formats. Early unified formulations such as One for All (OFA) move in this direction by seeking a single graph model for multiple graph classification settings through unified task representations and prompting \citep{Liu2023OneFA}. More recent GFM work makes this agenda explicit. GraphFM emphasizes scalable multi-graph, multi-task pretraining over a large collection of graph datasets, focusing on transferable node-level prediction \citep{Lachi2024GraphFMAS}. AnyGraph targets cross-domain generalization, heterogeneity handling, and fast adaptation across diverse graph distributions \citep{Xia2024AnyGraphGF}. LangGFM explores whether a large language model itself can serve as a universal graph learner across benchmark graph tasks \citep{Lin2024LangGFMAL}.

These works are important because they establish the mainstream GFM agenda: learning a general graph representation that works across graphs, datasets, and standard prediction tasks. Our setting, however, is different. GOFM is better understood as a specialized adaptation of the foundation-model idea to graph optimization in operations research. We focus on a setting where the underlying weighted infrastructure network remains fixed while optimization instances change over time. This places less emphasis on cross-graph transfer, but more emphasis on learning a reusable graph-specific structural prior that supports downstream feasible decision construction under task-specific constraints. Our contribution is therefore complementary to the GFM literature. Mainstream GFMs prioritize cross-graph generalization for basic tasks, while GOFM develops a graph-native prior for feasibility-critical optimization on persistent networks.

\subsection{Research gap and our positioning}

\begin{table}[ht]
\centering
\vspace{-8pt}
\caption{Positioning of GOFM relative to related research streams.}
\label{tab:positioning_matrix}
\footnotesize
\setlength{\tabcolsep}{3pt}
\renewcommand{\arraystretch}{1.18}

\begin{tabularx}{\linewidth}{@{}
    p{0.22\linewidth}
    >{\centering\arraybackslash}p{0.08\linewidth}
    >{\centering\arraybackslash}p{0.07\linewidth}
    >{\centering\arraybackslash}p{0.07\linewidth}
    >{\scriptsize\arraybackslash}X
@{}}
\toprule
\textbf{Research stream} &
\textbf{Sparse \newline feasibility} &
\textbf{Cross-task \newline reuse} &
\textbf{Single-graph \newline focus} &
\textbf{Representative references / notes} \\
\midrule

Neural combinatorial optimization / routing &
$\triangle$ &
$\times$ &
$\times$ &
Task-specific policies trained on complete-graph Euclidean instances; metric closure needed for sparse networks
\citep{Vinyals2015PointerN,Bello2016NeuralCO,Kool2018AttentionLT,Kwon2020POMOPO}. \\[4pt]

Graph representation learning &
$\times$ &
$\triangle$ &
$\triangle$ &
Reusable node embeddings for prediction and analysis, but no feasibility-constrained optimization or generative decoding
\citep{Kipf2016GCN,Grover2016node2vec}. \\[4pt]

LLM-based optimization &
$\triangle$ &
$\triangle$ &
$\times$ &
Flexible prompting and tool-use pipelines, but solution feasibility degrades rapidly as graph size grows
\citep{Yang2023LargeLM,Zhang2025ORLLMAgentAM}. \\[4pt]

Graph foundation models &
$\times$ &
$\triangle$ &
$\times$ &
Cross-graph pretraining for node/graph prediction, not for feasibility-constrained optimization on a fixed network
\citep{Liu2023OneFA,Lachi2024GraphFMAS,Xia2024AnyGraphGF,Lin2024LangGFMAL}. \\[2pt]

\midrule
\textbf{GOFM (ours)} &
$\checkmark$ &
$\checkmark$ &
$\checkmark$ &
\textbf{\scriptsize Self-supervised structural prior on a fixed graph, combined with constrained decoding across multiple optimization tasks.} \\

\bottomrule
\end{tabularx}

\vspace{0.4em}
\begin{minipage}{0.98\linewidth}
\scriptsize
\textit{Note.} $\checkmark$ = full support \quad
$\triangle$ = partial or limited \quad
$\times$ = not supported or not the primary design focus.
Assessments reflect the primary design emphasis of each stream in the context of distance-based optimization on sparse road networks. 
\end{minipage}
\vspace{-10pt}
\end{table}

Table~\ref{tab:positioning_matrix} summarizes how existing streams address the requirements of the single-graph, multi-task regime and positions our GOFM framework relative to each. Four patterns emerge from the review above. First, neural combinatorial optimization methods learn task-specific solvers on synthetic complete-graph distributions. They achieve strong performance in those settings but lack the topological knowledge needed for feasible, high-quality decisions on sparse infrastructure graphs, and their single-task training prevents knowledge transfer across problem families. Second, graph representation learning produces reusable embeddings, but these are optimized for prediction rather than for constructing feasible solutions under hard combinatorial constraints. Third, LLM-based approaches offer flexibility through natural-language interfaces and tool use, but their language-token representations are not aligned with graph structure, and feasibility collapses as problem size grows. Fourth, graph foundation models emphasize cross-graph generalization, not feasibility-constrained optimization or the construction of paths, tours, or route sets under hard combinatorial constraints.


Taken together, these streams leave an important gap: a graph-native framework tailored to sparse, weighted, fixed-network settings in which the same infrastructure supports multiple downstream optimization tasks. GOFM is positioned for this regime. It combines graph-native pretraining with constraint-aware decoding so that a single learned backbone can be reused across multiple tasks on the same graph, without task-specific retraining.

\section{Preliminaries}
\label{sec:preliminary}
This section introduces the formal setup and conceptual ingredients that underpin our Graph Optimization Foundation Model (GOFM). Specifically, we instantiate five representative graph optimization tasks that span the main problem classes studied in this paper and serve to demonstrate the breadth of GOFM (Section~\ref{sec:prelim-go}). 
Then, we trace the line of reasoning that led us from language-model pretraining to a graph-native foundation model (Section~\ref{sec:prelim-ss}).


\subsection{Graph Optimization Problems}
\label{sec:prelim-go}

We consider optimization problems defined on a weighted graph $G=(V,E,w)$, where $V$ is the node set, $E$ is the edge set, and $w_{ij}>0$ denotes the cost/weight associated with edge $(i,j)\in E$. A graph optimization problem seeks a structured decision $x$ induced by $G$ (e.g., a path, a tour, a set of routes, a partition, or a seed set) that optimizes an objective under feasibility constraints:
\begin{equation}
\min_{x \in \mathcal{X}(G)} f(x;G,\theta),
\label{eq:go}
\end{equation}
where $\mathcal{X}(G)$ specifies the feasible solution family on $G$ and $\theta$ denotes problem-specific inputs (e.g., terminals, required nodes, demands, budgets, or diffusion parameters).
This formulation provides the common language we use when presenting GOFM in subsequent sections.

\begin{figure*}[!t]
    \centering

    \newcommand{\RowOneH}{4.5cm}  
    \newcommand{\RowTwoH}{5.0cm}  
    \newcommand{\SubLabGap}{0pt}

    \begin{minipage}[t]{0.045\textwidth}
        \centering
        \vspace*{-88pt} 
        \rotatebox{90}{\scriptsize\textbf{Road Network}}
    \end{minipage}
    \hfill
    \begin{minipage}[t]{0.29\textwidth}
        \centering
        \includegraphics[height=\RowOneH,keepaspectratio]{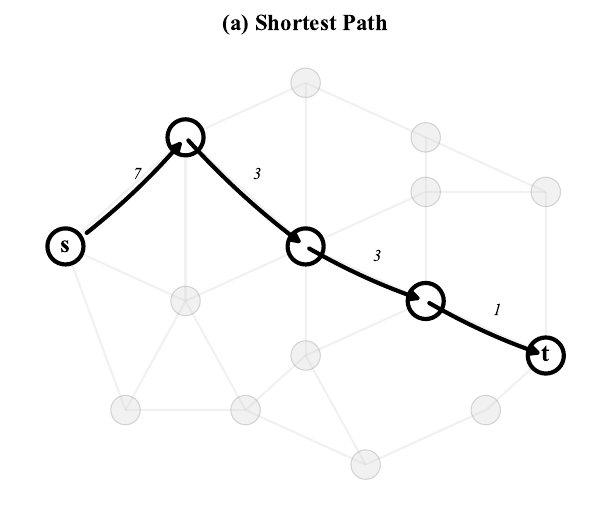}\par
    \end{minipage}
    \hfill
    \begin{minipage}[t]{0.29\textwidth}
        \centering
        \includegraphics[height=\RowOneH,keepaspectratio]{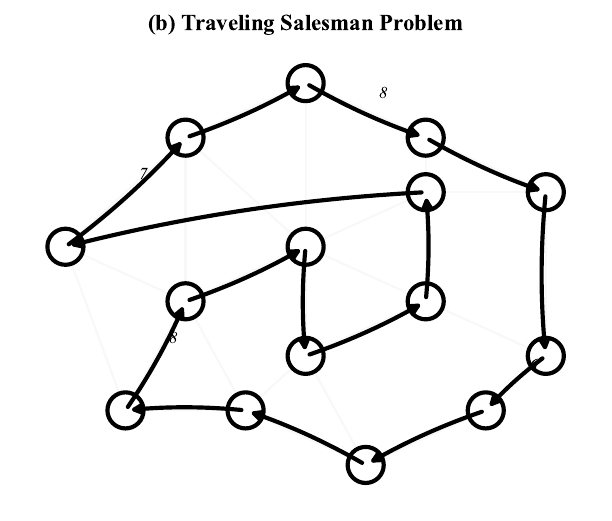}\par
    \end{minipage}
    \hfill
    \begin{minipage}[t]{0.29\textwidth}
        \centering
        \includegraphics[height=\RowOneH,keepaspectratio]{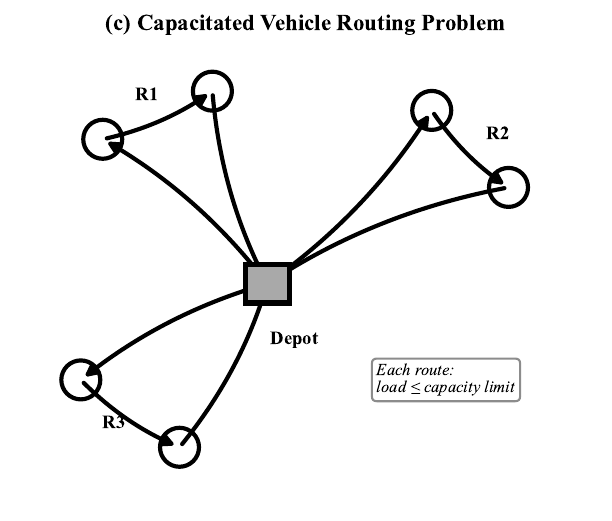}\par
    \end{minipage}

    \begin{minipage}[t]{0.045\textwidth}
        \centering
        \vspace*{-98pt} 
        \rotatebox{90}{\scriptsize\textbf{Social Network}}
    \end{minipage}
    \hfill
    \begin{minipage}[t]{0.44\textwidth}
        \centering
        \includegraphics[height=\RowTwoH,keepaspectratio]{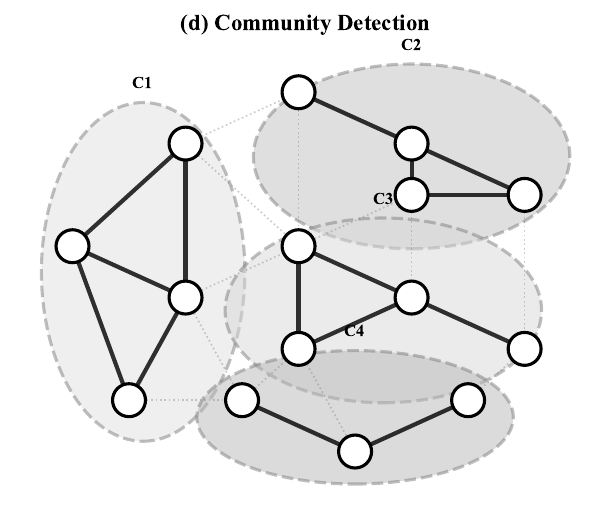}\par
    \end{minipage}
    \hfill
    \begin{minipage}[t]{0.44\textwidth}
        \centering
        \includegraphics[height=\RowTwoH,keepaspectratio]{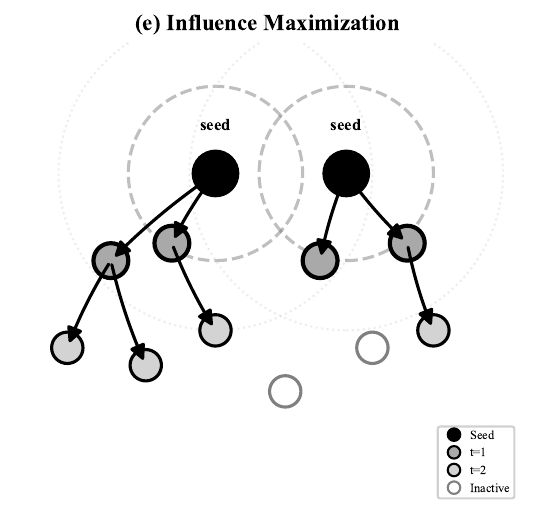}\par
    \end{minipage}

    \vspace{-2pt}
    \caption{Schematic illustrations of graph optimization problems: shortest path (SP), traveling salesman problem (TSP), capacitated vehicle routing problem (CVRP), community detection (CD), and influence maximization (IM).}
    \vspace{-4pt}
    \label{fig:prelim-problems}
\end{figure*}

Figure~\ref{fig:prelim-problems} summarizes five classical problem classes that will serve as running examples. 
We use $i$ to index nodes in the graph ($v_i\in V$), and use $j$ to index positions in an ordered solution sequence (e.g., $v_{(j)}$ denotes the $j$-th visited node in a path/tour).
When a solution contains multiple routes, we additionally use ``$\mid$'' as a route-separator token.
Below we make this explicit for each problem.

\paragraph{(a) Shortest Path (SP).}
\label{sec:prelim-sp}

Given an origin--destination (OD) pair $(o,d)\in V\times V$, the shortest path problem seeks a minimum-cost $o$--$d$ path on $G$.
Let $V=\{v_1,\ldots,v_n\}$ with $n=|V|$. A feasible path can be written as
\[
p=(v_{(1)} \to v_{(2)} \to \cdots \to v_{(m)}), \qquad
c(p)=\sum_{j=1}^{m-1} w\!\big(v_{(j)},v_{(j+1)}\big),
\]
with $v_{(1)}=o$, $v_{(m)}=d$, and $(v_{(j)},v_{(j+1)})\in E$ for all $j=1,\ldots,m-1$.
SP minimizes $c(p)$ over all feasible $o$--$d$ paths. SP is polynomial-time solvable and is a basic primitive underlying many routing and network optimization models.

\paragraph{(b) Traveling Salesman Problem (TSP).}
\label{sec:prelim-tsp}

Given a weighted graph $G=(V,E,w)$, the traveling salesman problem (TSP) seeks a minimum-cost cycle that visits every node in $V$ exactly once and returns to the start. In distance-based settings, we work with the graph-induced metric $d_G(\cdot,\cdot)$, where the distance between two nodes is the length of the shortest path in the underlying graph rather than their Euclidean separation.
Let $V=\{v_1,\ldots,v_n\}$ with $n=|V|$. A feasible tour can be written as
\[
\tau=(v_{(1)} \to v_{(2)} \to \cdots \to v_{(n)} \to v_{(1)}), \qquad
c(\tau)=\sum_{j=1}^{n} d_G\!\bigl(v_{(j)},v_{(j+1)}\bigr), \quad v_{(n+1)}:=v_{(1)},
\]
where $\{v_{(1)},\ldots,v_{(n)}\}=V$, so each node appears exactly once. TSP minimizes $c(\tau)$ over all feasible tours. 
TSP is NP-hard and serves as a canonical routing problem in which the key challenge is to determine a globally efficient visiting order under graph-based distances.
As the canonical single-tour routing problem, TSP also serves as the conceptual baseline for a broader tour problem family of optimization problems that modify the visiting requirements, cost structure, or resource constraints. Several such variants will be considered in our experiments (Section~\ref{sec:data}).

\paragraph{(c) Capacitated Vehicle Routing Problem (CVRP).}
\label{sec:prelim-cvrp}
CVRP extends TSP problems by introducing multiple vehicles and capacity constraints. Given a depot $o\in V$, a set of customers $R\subseteq V$ with demands $\{q_u\}_{u\in R}$, vehicle capacity $Q$, and an upper bound $m$ on the number of vehicles, the goal is to construct at most $m$ depot-rooted routes that collectively serve all customers, satisfy route-wise capacity constraints, and minimize total travel cost \citep{Laporte2009FiftyYO,Toth2014VehicleRP}.

We measure travel cost using the same graph-induced metric $d_G(\cdot,\cdot)$. A feasible solution is a set of routes
$\mathcal{R}=\{R_1,\ldots,R_K\}$ with $K\le m$, where
\[
R_k=(o \to v^{k}_{(1)} \to v^{k}_{(2)} \to \cdots \to v^{k}_{(m_k)} \to o), \qquad
\sum_{u\in R_k} q_u \le Q,\quad k=1,\ldots,K.
\]
Each customer in $R$ must appear in exactly one route in $\mathcal{R}$. The objective is
to minimize
\[
\sum_{k=1}^{K}\sum_{j=0}^{m_k} d_G\!\bigl(v^{k}_{(j)},v^{k}_{(j+1)}\bigr),
\qquad
v^{k}_{(0)}:=o,\; v^{k}_{(m_k+1)}:=o.
\]
Compared with TSP, CVRP couples decisions across routes and imposes hard resource constraints, substantially increasing modeling and algorithmic complexity.

\paragraph{(d) Other graph optimization problems (social networks) - Community Detection (CD).}
Beyond routing problems on road networks, we also consider representative social-network optimization tasks.
One established example is \emph{community detection} (CD), which seeks a partition of $V$ into $K$ communities
$\Pi=\{C_1,\ldots,C_K\}$ such that connections are dense within communities and sparse across communities.
Equivalently, under an ordered node list $V=\{v_1,\ldots,v_n\}$, a feasible solution can be represented by a label vector
$y=(y_1,\ldots,y_n)$ with $y_i\in\{1,\ldots,K\}$, where $y_i$ assigns node $v_i$ to community $y_i$.
Typical objectives minimize cross-community connectivity or maximize modularity.

\paragraph{Influence Maximization (IM).} This problem aims to select a seed set $S\subseteq V$ with $|S|\le k$
to maximize the expected diffusion reach over the network under a stochastic propagation model (e.g., independent cascade or linear threshold):
\[
\max_{S\subseteq V,\ |S|\le k}\ \sigma_G(S)\ :=\ \mathbb{E}\big[|A(S)|\big],
\]
where $A(S)$ denotes the (random) activated set generated by the diffusion process starting from $S$. Unlike routing objectives that are additive over edge costs, the IM objective is induced by a dynamic process on the graph. In Section~\ref{sec:data}, we detail the datasets and problems for our experiments.

All the problem classes share a common characteristic: the decision is a structured object (path, tour, route set, partition, or seed set) that must respect the connectivity and cost structure of a fixed underlying graph.
This observation is the starting point for GOFM. If a model can internalize the topological and metric regularities of $G$ in a task-agnostic way, the same learned representation should be useful across all these problem classes. Next, we explain how we arrived at this idea by drawing on the self-supervised pretraining principles that underpin modern language models.

\subsection{From Language Self-Supervision to GOFM}
\label{sec:prelim-ss}
The goal of this subsection is to trace the line of reasoning that bridges self-supervised learning in natural language to graph optimization. We begin with the observation that made Word2Vec work, show how Node2Vec adapted it to graphs, and then explain the additional step needed to move from learning static node embeddings to learning a generative structural prior for optimization.

\subsubsection{From Word2Vec to Node2Vec.}
\label{sec:prelim-w2v-n2v}


The working mechanism of modern language models follows a simple recipe: build a large corpus, pretrain on it with a self-supervised objective, and reuse the learned representations at inference time. The reason this recipe works is that the structure of the data itself provides supervision at scale. In language, raw text is naturally a sequence corpus, and objectives such as context prediction or masked reconstruction force the model to internalize which tokens tend to co-occur and which completions are plausible in a given context. In graphs, the raw object is not sequential, but a sequence corpus can be constructed by sampling graph-consistent trajectories. This is exactly the insight behind walk-based methods such as Node2Vec \citep{Grover2016node2vec}: nodes serve as ``tokens'' and walks serve as ``sentences,'' enabling self-supervised learning without task-specific labels, as reviewed in Section~\ref{sec:lr-graph-repr}.

GOFM adopts a similar transfer but changes the target of learning. Rather than learning static embeddings primarily for downstream prediction or clustering, we aim to learn a graph-conditioned structural prior that is directly useful for distance-based decisions on weighted networks. Intuitively, the prior should capture the ``graph grammar'' of plausible traversals and subpath composition under the weighted geometry: given a partially observed trajectory (anchors on both sides), which intermediate nodes or substructures are compatible with completing that traversal on the network? This shift from \emph{static} representation to generative structural prior is our first key novelty.

\subsubsection{Transformer Pretraining Objectives.}
\label{sec:prelim-bert-gpt}

While Word2Vec and Node2Vec learn \emph{static} embeddings from local co-occurrence, the next step in the timeline is the shift to contextual representations learned by deep sequence models. Transformer architectures enable scalable sequence modeling via self-attention, and two pretraining objectives have become widely used.

\paragraph{Masked language modeling (BERT-style).} BERT \citep{BERT} learns bidirectional contextual representations by masking a subset of tokens and training the model to reconstruct the masked content from its surrounding context. This denoising-style objective produces a general-purpose encoder that transfers well to downstream tasks through fine-tuning.

\paragraph{Autoregressive next-token prediction (GPT-style).}
GPT-style language models are trained to predict the next token given all preceding tokens \citep{chatgpt}. This objective yields a generative model that supports zero- and few-shot adaptation via prompting and can be further specialized through instruction tuning or preference optimization.

Both objectives share a common logic: large-scale sequential data provides abundant self-supervised training signal, and the resulting model captures reusable statistical structure that can be transferred to new tasks. The question is whether an analogous logic applies to graphs.

\subsubsection{Bridging Logic: From LLM Pretraining to GOFM.}
\label{sec:prelim-bridge}

\begin{figure}[t]
    \centering
    \includegraphics[width=0.92\linewidth]{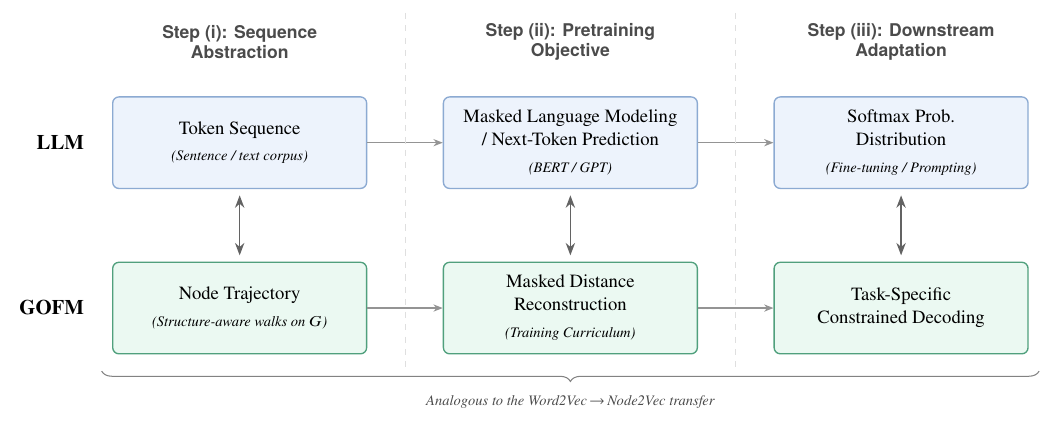}
    \vspace{-7pt}
    \caption{From language self-supervision to GOFM: a conceptual mapping.}
    \label{fig:llm_to_dgfm_mapping}
    \vspace{-6pt}
\end{figure}

Our goal is to build an analogous bridge from sequence foundation models to graph optimization. 
The conceptual mapping mirrors the Word2Vec$\rightarrow$Node2Vec transfer, but requires three deliberate steps (as shown in Figure~\ref{fig:llm_to_dgfm_mapping}):
(i) identify an appropriate \emph{sequence view} of graph structure,
(ii) define a scalable self-supervised objective that learns transferable priors, and
(iii) combine the learned prior with explicit task constraints during downstream decision making under task-specific requirements.

Concretely, Step~(i) constructs a graph-consistent trajectory corpus from $G$ via structure-aware random walks, as in Node2Vec. Step~(ii) is where our approach departs most clearly from prior graph representation methods: rather than training a context-prediction model that produces static embeddings, we pretrain a Transformer with masked trajectory reconstruction to obtain a generative structural prior over plausible traversals on the network. This prior captures how subpaths compose on the graph under weighted adjacency, which is precisely the structural knowledge needed for distance-based optimization. Lastly, Step~(iii) addresses a challenge that has no direct analogue in language modeling. In natural language processing, downstream tasks are typically handled by fine-tuning or prompting, neither of which involves hard combinatorial constraints. In graph optimization, however, the output must be a feasible solution in $\mathcal{X}(G)$ of Equation~\eqref{eq:go}, meaning that any decoded path or tour must respect connectivity, capacity, and other structural constraints. GOFM therefore separates the learned prior from the constraint enforcement: the prior proposes structurally compatible candidates, and a task-specific decoding layer ensures that the final solution satisfies all required constraints. Section~\ref{sec:gofm} details the GOFM pretraining objective and the task-specific decoding procedures.

\section{GOFM Framework}
\label{sec:gofm}
\begin{figure*}[!t]
  \centering
  \includegraphics[width=0.95\textwidth]{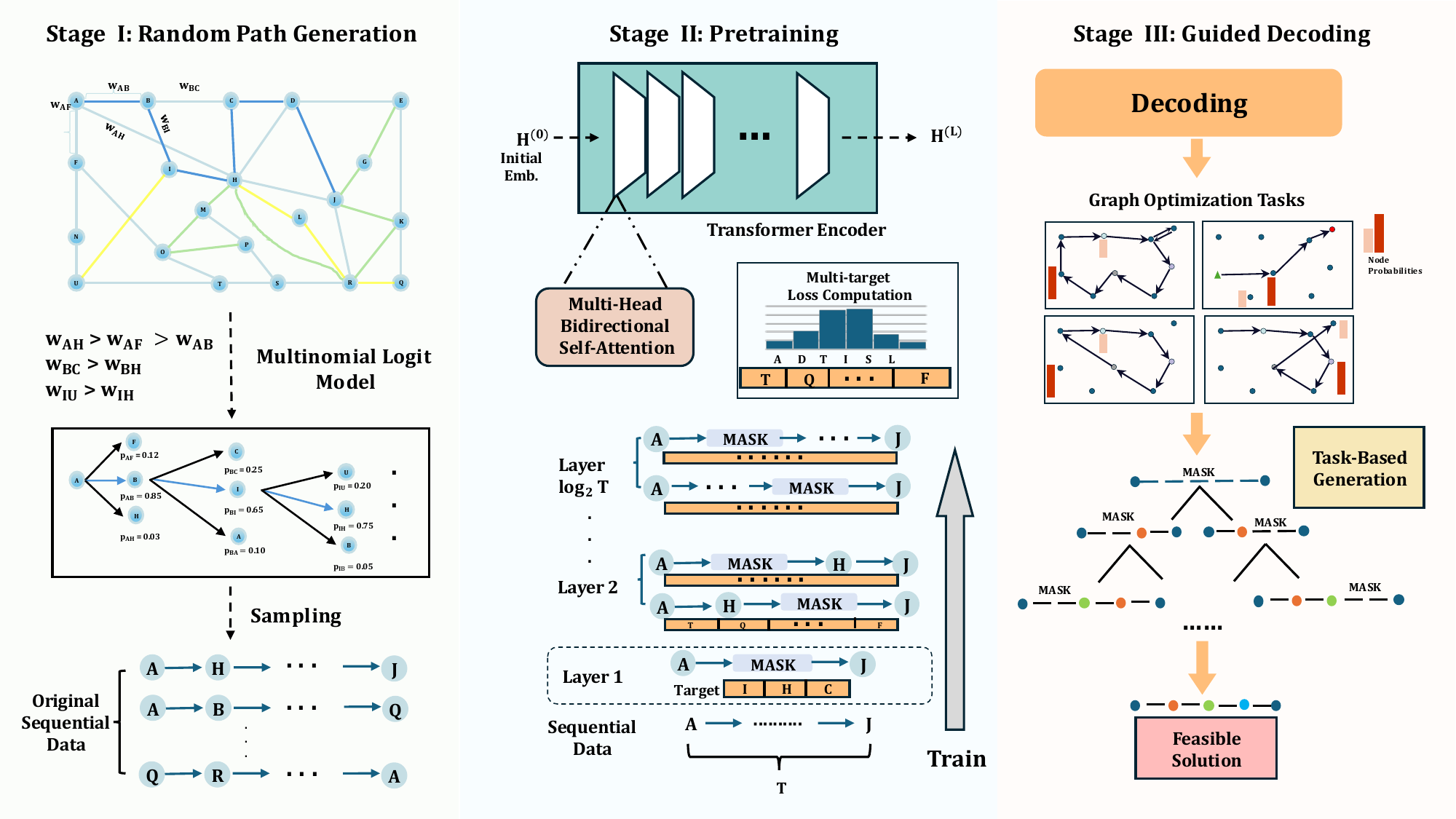}
  \caption{
  The overall framework of our proposed GOFM.
  }
  \begin{minipage}{\textwidth}
    \vspace{.1in} \scriptsize
    Note. Distance-biased random walks generate graph-consistent training trajectories. Node and position embeddings are fed into a bidirectional Transformer encoder trained by multi-target reconstruction, and the learned prior is later combined with task-specific decoding for downstream graph optimization tasks.
    \end{minipage}
    \vspace{-10pt}
  \label{fig:framework}
\end{figure*}

In this section, we operationalize the conceptual mapping in Section~\ref{sec:prelim-ss} into a concrete Graph Optimization Foundation Model (GOFM) framework for graph optimization on a persistent weighted graph $G=(V,E,w)$. Our focus is \emph{single-graph, multi-task} generalization: the underlying network remains fixed, while task specifications vary over time. This setting is the primary deployment regime considered in our study and motivates a graph-native pretrain--decode pipeline rather than per-task retraining. The framework consists of three stages. Figure~\ref{fig:framework} provides an overview, and the remainder of this section details each stage.


\noindent\textbf{Stage I: Random Path Generation (Section~\ref{sec:RW/Path generation}).}
The graph itself provides abundant self-supervised data. We generate a large corpus of graph-consistent trajectories using a distance-biased, Node2Vec-style random walk mechanism. This corpus is built to reflect both the topology of $G$ and the weighted geometry induced by $w$.

\smallskip
\noindent\textbf{Stage II: Pretraining (Section~\ref{sec:pretraining}).}
We treat each trajectory as a ``sentence'' over node ``tokens'' and pretrain a bidirectional Transformer to reconstruct masked portions of trajectories from their context. The outcome is a task-agnostic structural prior $\pi_\theta(\cdot \mid G)$ that captures the ``graph grammar'' of plausible traversals and subpath compatibility on the fixed network.

\smallskip
\noindent\textbf{Stage III: Guided Decoding (Section~\ref{sec:decoding}).}
Given a downstream specification $s$, we decode solutions without retraining by combining the learned prior with explicit task constraints and objectives. Decoding is restricted to the feasible set $\mathcal{F}(G,s)$, while $\pi_\theta(\cdot \mid G)$ serves as a guidance signal for constructive decisions or candidate refinements.

\subsection{Stage I: Random Path Generation}
\label{sec:RW/Path generation}

A prerequisite for self-supervised pretraining is an abundant source of data. In language, this role is naturally played by the abundance of large-scale text corpora. For a network such as a city-scale road graph, however, the learning signal must come from the structure itself. Our key design choice is to convert the weighted graph into sequential trajectories that satisfy three properties: (i) they respect graph connectivity, so the model never observes illegal transitions, (ii) they cover diverse regions and motifs of the graph, and (iii) they reflect the metric preferences induced by edge weights, so that shorter edges are statistically favored. This conversion yields a graph corpus analogous to text, where nodes play the role of words and random-walk trajectories play the role of sentences.

For downstream distance-based optimization on road networks, the structural prior should internalize two complementary signals: \emph{topology} (adjacency, bridges, and community boundaries) and \emph{geometry} (short links are more natural under shortest-path metrics). Purely unbiased random walks over-emphasize high-degree regions and do not encode a preference for shorter edges. Conversely, a greedy shortest-edge policy collapses exploration and yields a narrow corpus. We therefore need a mechanism that is graph-consistent by design but can move between local exploitation and global exploration, while injecting a controllable bias toward low-cost edges.

\paragraph{Node2Vec-style biased walks with distance-aware logits.}
We adopt a Node2Vec-style second-order biased random walk \citep{Grover2016node2vec} because it provides exactly this balance through two parameters $(p,q)$ that control local versus global exploration, and we extend it with distance-aware logits to incorporate edge weights. The parameter $p$ discourages backtracking (preventing degenerate oscillations), while $q$ controls the local--global trade-off (BFS-like vs.\ DFS-like).

Formally, starting from node $i=v_t$, the probability of moving to a neighbor $j$ given the previous node $r=v_{t-1}$ is
\begin{equation}
\Pr(v_{t+1}=j \mid v_t=i, v_{t-1}=r) = \frac{\exp\!\big(-\beta \cdot w_{ij}\cdot b_{p,q}(i,j,r)\big)}{\sum_{k\in\mathcal{N}(i)} \exp\!\big(-\beta \cdot w_{ik}\cdot b_{p,q}(i,k,r)\big)},
\label{eq:transition}
\end{equation}
where $w_{ij}$ is the edge weight, $\beta>0$ is a softmax scaling factor controlling the sharpness of the distribution, and $b_{p,q}(i,j,r)$ denotes the Node2Vec bias term. This construction produces trajectories whose empirical frequency reflects both topology and geometry, which are exactly the two ingredients we want the structural prior to learn.

\paragraph{Weight normalization and temperature.}
\label{sec:weight_norm_temp}
Raw edge weights can vary by orders of magnitude across a road network. If used directly in the softmax transition rule of Equation~\eqref{eq:transition}, the resulting distribution may become overly peaked (collapsing toward near-greedy choices) or, after rescaling, become nearly uniform (washing out metric information). We address this with two steps: (i) normalizing edge costs to remove graph-specific scale effects, and (ii) introducing an inverse-temperature parameter $\beta$ to control how strongly the walk prefers low-cost edges.

Specifically, we use the normalized cost
\begin{equation}
\tilde w_{ij}
=
\frac{w_{ij}}{\mathrm{median}\{w_{ab}:(a,b)\in E\}+\epsilon},
\label{eq:weight_normalize}
\end{equation}
where $\epsilon>0$ is a small constant for numerical stability.
In implementation, we replace $w_{ij}$ by $\tilde w_{ij}$ in the softmax of Equation~\eqref{eq:transition}.
Larger $\beta$ makes transitions more cost-sensitive (favoring shorter edges), while smaller $\beta$ yields more exploratory walks. In practice, we select $\beta$ so that, for a typical node, multiple neighbors retain meaningful probability mass. This prevents the corpus from collapsing into greedy low-cost traversals while still encoding metric preference, and it makes the hyperparameter comparable across networks of different scale.

\paragraph{Corpus generation protocol.}
We generate a corpus by sampling multiple trajectories from diverse starting nodes. We sample $M$ walks per node (or per a subset of nodes) with walk length $L$,
yielding a corpus size of approximately $|V|\cdot M \cdot L$ tokens.
To avoid over-representing high-degree regions and to improve coverage, we use one of two starting-node strategies:
(i) \emph{uniform starts} over $V$, or (ii) \emph{degree-tempered starts} with $\Pr(v_0=u)\propto \deg(u)^\alpha$ for $\alpha\in(0,1)$.
The latter balances uniform coverage with the practical importance of high-traffic junctions on road networks.

Each walk is recorded as an ordered node sequence $\mathbf{s} =[v_1, v_2, \ldots, v_L]$,
which can be treated as a ``sentence'' for pretraining.
We optionally augment the sequence with lightweight structural tokens
but keep the vocabulary minimal to emphasize transferable structural regularities.

The resulting corpus has two desirable properties.
First, it is \emph{graph-consistent}: each consecutive node pair is an actual edge, so the model never observes illegal local moves.
Second, it is \emph{metric-informative}: due to the $\exp(-\beta \tilde w_{ij})$ term, shorter edges are statistically over-represented, which aligns the pretraining signal with shortest-path-based objectives. At the same time, the Node2Vec bias prevents the corpus from collapsing to trivial local patterns and preserves exploration across communities. This matters especially on road networks, where optimal tours and routes depend on both local geometry and global connectivity.

\medskip
\noindent With the trajectory corpus in hand, the next question is how to distill its structural regularities into a learnable model. We address this through a progressive masked reconstruction objective.

\subsection{Stage II: Pretraining}
\label{sec:pretraining}

Stage II takes the trajectory corpus from Stage I and distills its structural regularities into a learnable model. This involves two coupled design decisions: a training objective that specifies what the model should learn (Section~\ref{sec:prog-objective}), and an encoder architecture that determines how the model processes masked trajectories to produce node predictions (Section~\ref{sec:encoder}). 

\subsubsection{Progressive Training.}
\label{sec:prog-objective}

The self-supervised learning problem is to predict masked portions of a trajectory from the observed context, forcing the model to internalize both global connectivity and local consistency. This is analogous to masked language modeling in BERT \citep{BERT}, where predicting masked tokens from surrounding context teaches the model reusable linguistic structure. On graphs, the analogue of a token is a node in a trajectory, and the analogue of context is the observed portion of a walk.

A direct masked prediction formulation on long trajectories can be unstable, however. Early in training, the model has limited ability to infer long-range structure, and masking many interior nodes at once may encourage shortcut heuristics (e.g., predicting the most frequent node) that do not generalize. To address this, we adopt a \emph{progressive training curriculum} that moves from coarse, global reconstruction to finer, local reconstruction. The idea is organized as a balanced refinement process: each step introduces a small number of anchors that split a long masked gap into shorter sub-gaps, so the typical reconstruction span decreases by a roughly constant factor.

This design yields two benefits.
First, it provides a principled coarse-to-fine signal: early levels force the model to learn global compatibility (which regions can plausibly connect), while later levels emphasize local consistency (how short subpaths should be completed between nearby anchors).
Second, it avoids redundant supervision and yields a logarithmic depth schedule.
Since each refinement roughly halves the remaining segment length, after $O(\log T)$ refinements (for a trajectory of length $T$) additional splits contribute diminishing new information but increase the number of training queries.
This coarse-to-fine schedule follows the same intuition as the balanced binary-tree insertion order used in the Insertion Transformer \citep{Stern2019InsertionTF}.

At the first level ($k=1$), only the endpoints are observed while the entire interior is masked,
\begin{equation}
    \ell^{1} = [v_1, \text{MASK}, v_T], \quad\mathcal{Y} = \{v_2, \ldots, v_{T-1}\},
\end{equation}
where $\ell^{1}$ is the level-1 masked query and $\mathcal{Y}$ is the admissible target set.
This stage encourages the model to learn a global prior over which nodes are compatible with the $(v_1,v_T)$ context.

At higher levels ($k>1$), we reveal $k-1$ interior nodes
$\{a_1,\dots,a_{k-1}\}$ as anchors, so that the reconstruction task
breaks into shorter subproblems between consecutive anchors. This
refinement enforces local consistency while remaining aligned with the
same global trajectory. Let the original trajectory be denoted by
$\mathbf{s}=(v_1,\dots,v_T)$, and define boundary anchors
$a_0:=v_1$ and $a_k:=v_T$. For each adjacent anchor pair
$(a_j,a_{j+1})$ with $j=0,1,\dots,k-1$, we construct one masked sequence with admissible target set
\(
\mathcal{Y}^{k}_{j}
=
\{\, u \in V(\mathbf{s}) \mid a_j \prec_{\mathbf{s}} u \prec_{\mathbf{s}} a_{j+1} \,\},
\)
where $V(\mathbf{s})$ denotes the set of nodes appearing in trajectory
$\mathbf{s}$, and $\prec_{\mathbf{s}}$ denotes their ordering along
$\mathbf{s}$. The masked sequence is
\begin{equation}
\ell^{k}_{j} =
[v_1, a_1, \ldots, a_j, \text{MASK}, a_{j+1}, \ldots, v_T].
\end{equation}
Hence, the level-$k$ training set is
\begin{equation}
\mathcal{L}^{k}
=
\{\, (\ell^{k}_{j},\, \mathcal{Y}^{k}_{j}) \mid j=0,\dots,k-1 \,\}.
\end{equation}

\paragraph{Multi-target loss.}
A further modeling consideration is that graph trajectories admit multiple valid completions even under the same context. Between two anchors, several nodes may all be consistent with the observed walk segment, especially in dense local neighborhoods. Standard single-target cross-entropy would penalize the model for assigning probability to any valid completion other than one designated ``correct'' answer, which is undesirable when multiple completions are equally legitimate. To handle this ambiguity, we adopt a multi-target loss that aggregates probability mass over the entire admissible target set. For a masked query $x$ with admissible target set $\mathcal{Y}_s$, let $z\in\mathbb{R}^{|V|}$ be the logits and $P=\softmax(z)$. The loss is
\begin{equation}
\ell_{\text{MT}}(s) = -\log \sum_{y\in\mathcal{Y}_s} P_y,
\end{equation}
which equals the negative log-probability that the model places on any valid target. The final objective averages this loss across a mini-batch $\mathcal{B}$:
\begin{equation}
\mathcal{L}_{\text{prog}}
= \tfrac{1}{|\mathcal{B}|}\sum_{x\in \mathcal{B}} \alpha_{\ell(x)} \,\ell_{\text{MT}}(x),
\end{equation}
where $x$ indexes a query, $\ell(x)$ is its granularity level, and $\alpha_\ell > 0$ is a normalized level-dependent weight; unless otherwise stated, we use the same default weights across experiments.

\begin{algorithm}[t]
\footnotesize
\caption{Progressive Training Construction from Random Walks}
\label{alg:prog}
\begin{algorithmic}[1]
\REQUIRE Random walk $\mathbf{v}=[v_1,\dots,v_T]$ on graph $G$, level bounds $L_{min}$,$L_{\max}$.
\ENSURE Training set $\mathcal{D}$ of masked-prediction samples.

\STATE Initialize $\mathcal{D}\gets\varnothing$, interior nodes $\mathcal{V}_{\mathrm{int}}=\{v_2,\dots,v_{T-1}\}$.
\STATE Determine number of levels $\ell_{\max}=\min(L_{\max},\max(L_{\min},\lfloor \log_2 T\rfloor))$.

\STATE \textbf{Level 1 (global prior).}
Construct $\mathbf{x}=[v_1,\,\textsc{Mask},\,v_T]$, with admissible targets $\mathcal{Y}=\mathcal{V}_{\mathrm{int}}$.
Add $(\mathbf{x},\mathcal{Y},\ell{=}1)$ to $\mathcal{D}$.
\STATE Initialize anchor set $C^{(1)}=[v_1,v_T]$. \hfill // $C^{(\ell)}$: anchor set at level $\ell$

\FOR{$\ell=2$ \TO $\ell_{\max}$}
    \STATE Select a gap $(a,b)$ from $C^{(\ell-1)}$ and sample anchor $u$ from $\textsc{Between}(a,b\mid \mathbf{v})$ (if empty, sample from $\mathcal{V}_{\mathrm{int}}$).
    \STATE Insert $u$ into $C^{(\ell-1)}$ to form $C^{(\ell)}$. \hfill // refine anchors
    \FORALL{gaps $(a',b')$ in $C^{(\ell)}$}
        \STATE Form query $\mathbf{x}$ by inserting \textsc{Mask} between $(a',b')$.
        \STATE Admissible targets $\mathcal{Y}=\textsc{Between}(a',b'\mid \mathbf{v})$.
        \STATE Add $(\mathbf{x},\mathcal{Y},\ell)$ to $\mathcal{D}$.
    \ENDFOR
    \STATE Optionally mask additional random interior positions of $C^{(\ell)}$ and generate samples analogously.
\ENDFOR
\STATE \textbf{return} $\mathcal{D}$
\end{algorithmic}
\end{algorithm}

Intuitively, earlier levels emphasize global compatibility while later levels emphasize local refinement. This structure balances these two pressures while keeping the number of refinement levels logarithmic in trajectory length. Algorithm~\ref{alg:prog} summarizes the full progressive training construction. We now turn to the model architecture that implements this objective.

\subsubsection{Transformer Encoder Architecture.}
\label{sec:encoder}

The choice of model architecture is driven by the structure of the pretraining task. In distance-based routing on sparse graphs, many supervision signals are inherently \emph{two-sided}. Consider a shortest-path trajectory that is only partially observed: when an interior segment is masked, predicting a missing node is not a left-to-right completion problem. It is a bridge problem, jointly constrained by the origin-side prefix and the destination-side suffix. Whether an interior node is feasible and plausible depends on its compatibility with both sides of the observed trajectory on the road network.

This observation motivates an encoder-only, \emph{bidirectional} attention architecture in the spirit of BERT-style masked reconstruction. Each position attends to the entire observed trajectory so that evidence from both earlier and later positions can inform the reconstruction of masked segments. A decoder that processes the trajectory strictly from left to right would artificially restrict the information available at each position and weaken the alignment between the objective of reconstructing masked segments from both directions and the model architecture.

Because our trajectories can be long, optimization stability becomes a practical concern. We implement the encoder blocks using a \emph{pre-norm} Transformer, where layer normalization is applied inside each residual block before the attention and MLP sublayers. This choice follows \citet{Xiong2020OnLN}, who show that pre-norm placement yields better-behaved gradients at initialization compared to the original post-norm design, improving training stability for long sequences.

Formally, let $H\in\mathbb{R}^{T\times d}$ denote the hidden states at the input of a Transformer layer, where $T$ is the trajectory length and $d$ is the embedding dimension. Multi-head self-attention computes
\begin{equation}
\mathrm{MHA}(H) =
\Big[\,
\softmax\!\Big(\tfrac{(H W_m^Q)(H W_m^K)^\top}{\sqrt{d_h}} + M_{\text{pad}}\Big)(H W_m^V)
\Big]_{m=1}^{M} W^O ,
\end{equation}
where $M$ is the number of heads, $d_h=d/M$ is the head dimension,
$W_m^Q,W_m^K,W_m^V\in\mathbb{R}^{d\times d_h}$ are the learned projection matrices for queries, keys, and values,
and $W^O\in\mathbb{R}^{Md_h\times d}$ is the output projection.
Here $Q_m=HW_m^Q$, $K_m=HW_m^K$, and $V_m=HW_m^V$ are the per-head query, key, and value representations, and $[\cdot]_{m=1}^{M}$ denotes concatenation over heads. We do not impose causal masking. The only masking applied is for padding, so valid positions retain full mutual visibility. $M_{\text{pad}}\in\mathbb{R}^{T\times T}$ is an additive mask with $0$ for valid tokens and $-\infty$ for padding positions. This bidirectional design allows the model to use full-sequence context for predicting masked nodes, distilling the structural regularities of weighted graph traversals into the learned parameters.

\smallskip
\noindent Stages I and II produce a task-agnostic structural prior $\pi(\cdot \mid G)$ that captures how plausible traversals compose on the given network. The remaining question is how to convert this prior into feasible solutions for specific optimization problems. We address this next.

\subsection{Stage III: Guided Decoding}
\label{sec:decoding}

\paragraph{Structural prior: a transferable graph-dependent bias.}
Our pretraining produces a \emph{structural prior} $\pi(\cdot\mid G)$: a graph-conditioned distribution that summarizes contextually topology and geometry with a partially observed traversal on the weighted network $G=(V,E,w)$. This prior captures recurring structural regularities of feasible movement under the network metric, distilled from masked-trajectory reconstruction. Importantly, the prior is \textit{task-agnostic}: it depends only on $G$ and not on any particular optimization objective or feasibility rule. As a result, it is transferable. Once $\pi(\cdot\mid G)$ is learned, we can address different graph optimization problems by changing only the decoding logic, without retraining the model. Formally, for a task specification $s$ (e.g., terminals, required nodes, capacity), we output a feasible structured solution via
\begin{equation}
\widehat{\mathbf{Y}} \;=\; \mathrm{Decode}\big(\pi(\cdot \mid G),\, s\big).
\end{equation}

\paragraph{A unified decoding principle.}
Decoding maps the task-agnostic prior to task-feasible decisions using two components:
(i) \emph{proposals} from the learned prior $\pi(\cdot\mid G)$ and (ii) \emph{task-specific constraint and cost} specified by $s$. Given the current partial solution $S_t$, we query the model to obtain a distribution $P(\cdot\mid S_t,G)$ over candidate updates, restrict to a small candidate set $\mathcal{C}_t$ (e.g., top-$K$ proposals), and then select the update by task-specific scoring with feasibility checks:
\[
S_{t+1} \;=\; \Pi_s\!\Big(S_t \oplus \arg\min_{c\in\mathcal{C}_t}\Delta J(c;S_t,G,s)\Big).
\]
Here $\Delta J(c;S_t,G,s)$ denotes the incremental objective change caused by update $c$, and $\Pi_s(\cdot)$ enforces hard constraints by projection or rejection. This separation of proposal generation (learned) from constraint enforcement (explicit) is the key design principle that enables cross-task reuse.

We now describe how this principle specializes to different problem classes. It affects only the definition of $\Delta J$, $\Pi_s$, and the node-selection strategy. The pretrained prior remains unchanged.

\paragraph{Shortest path (SP) decoding.}
For SP with terminals $(o,d)$, the constraint structure is simple and admits polynomial-time exact algorithms that are fundamentally best-first (e.g., Dijkstra, $A^*$). In our setting, the structural prior provides an informative local signal for which intermediate nodes are compatible with the current left-right context.
Accordingly, SP decoding follows a \emph{greedy} mode that prioritizes the highest-confidence proposals (argmax within admissible candidates), while feasibility is ensured by projecting any non-edge jump onto the underlying sparse network via shortest-path connections. Complete SP decoding details are deferred to the E-Companion Section~\ref{ec:decoding}.

\paragraph{Tour and routing decoding: why least-confidence-first.}
For tour-family problems and CVRP, decoding must jointly decide which node to incorporate next and where to insert it in the partial structure. We adopt an insertion-style constructive heuristic because it maintains a coherent partial solution throughout construction and allows incremental cost evaluation at each step.

Within this framework, the order in which nodes are inserted matters. A natural but naive strategy is to insert the ``easiest'' (highest-confidence) node first. The problem with this greedy-easy-first approach is that it tends to consume the most flexible insertion positions early, leaving the structurally difficult nodes for later stages when the partial tour has solidified and only expensive positions remain. The result is often a late-stage detour that substantially increases total cost.

We therefore adopt the opposite strategy: at each step, we insert the \emph{least-confident}, or hardest-to-place node first. Nodes that are currently incompatible with most gaps in the partial tour tend to remain difficult as the structure solidifies. Inserting them early, while the partial solution still has flexibility, reduces the risk of expensive late insertions. This design choice can be understood as a form of regret control: by prioritizing structurally uncertain decisions early, we preserve more options for the easier nodes that follow.

\paragraph{Hybrid decoding for CVRP.}
A key advantage of a task-agnostic structural prior is that it can be plugged into established routing heuristics as a data-driven proposal mechanism, while the heuristic layer enforces feasibility and refine costs. Rather than replace classical methods, $\pi(\cdot\mid G)$ complements them: it biases construction toward graph-consistent structures, and standard operators handle the hard constraints. This yields competitive solutions with fast polynomial-time inference.

We illustrate this hybrid principle on the CVRP decoder. We first decode a single customer ordering (a ``giant tour'') over $R$ using the same insertion-style logic as in tour decoding, with $\pi(\cdot\mid G)$ guiding which customers are structurally compatible with the evolving partial order under the road-network metric. We then project this permutation into a capacity-feasible multi-route solution using a standard split dynamic program that partitions the sequence into at most $m$ routes subject to vehicle capacity $Q$ \citep{Vidal2020HybridGS}.
Segment costs are computed using the shortest-path metric, and any segment violating capacity is disallowed by the feasibility operator $\Pi_s(\cdot)$.
Finally, we optionally apply lightweight local improvement within the same projected metric (e.g., 2-opt within routes and relocate/swap/2-opt$^\star$ across routes).

\begin{figure}[!htbp]
    \centering
    \includegraphics[width=\linewidth]{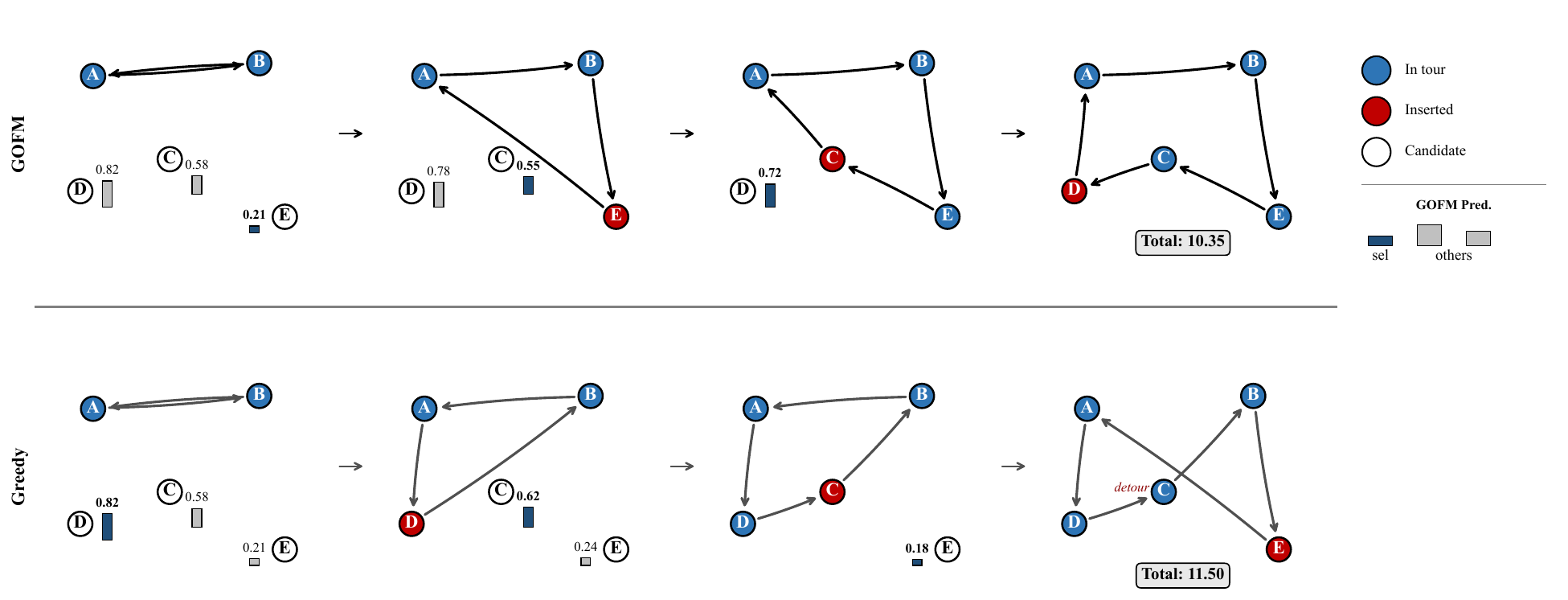}
    \caption{
    Illustration of least-confidence-first insertion.
    Top: GOFM decoding. Bottom: greedy decoding baseline.
    At each step, the pretrained prior provides scores for each uninserted node given the partial tour.
    }
    \label{fig:decoding-least-confidence}
    \vspace{-8pt}
\end{figure}

\paragraph{A concrete example: TSP family.}
To make the tour decoder concrete, we walk through the two-step procedure on a TSP-like problem with a required node set $R$. We maintain a partial tour as an anchor sequence $S_t=[a_1,\dots,a_L]$ and its gap set
$\mathcal{G}(S_t)=\{(a_\ell,a_{\ell+1})\}_{\ell=1}^{L-1}$ (and $(a_L,a_1)$ for cycle variants).
At each iteration, the prior is queried on each gap context to score how compatible an uninserted node is with being placed between that gap.

\emph{Step 1 (choose the next node to insert).}
For each unvisited node $u\in R\setminus S_t$, we compute a conservative difficulty score using the worst-gap compatibility:
\[
\kappa_t(u)\;=\;\min_{g\in \mathcal{G}(S_t)} P\big(u \mid \mathrm{ctx}(g),G\big),
\qquad
u_t^\star \;=\; \arg\min_{u\in R\setminus S_t}\kappa_t(u).
\]
This least-confidence-first rule selects the structurally hardest node while the tour is still flexible, reducing downstream insertion regret.

\emph{Step 2 (choose the insertion position).}
Given $u_t^\star$, we shortlist promising gaps by prior scores,
\[
\mathcal{G}_t(u_t^\star)
\;=\;
\operatorname{TopK}_{g\in \mathcal{G}(S_t)}
P(u_t^\star\mid \mathrm{ctx}(g),G).
\]
where $\operatorname{TopK}_{g\in \mathcal{G}(S_t)}(\cdot)$ returns the set of the top-$K$ gaps ranked by the indicated score and pick the position that minimizes incremental travel length under the projected metric $d_G(\cdot,\cdot)$:
\[
g_t^\star
\;=\;
\arg\min_{(a,b)\in \mathcal{G}_t(u_t^\star)}
\Delta J(u_t^\star;a,b),
\qquad
\Delta J(u;a,b)\;=\; d_G(a,u)+d_G(u,b)-d_G(a,b).
\]
We insert $u_t^\star$ into $S_t$ at $g_t^\star$ and repeat until all required nodes are included.
Figure~\ref{fig:decoding-least-confidence} shows how least-confidence-first avoids a late-stage detour compared to a greedy easy-first policy.

\section{Experiments}\label{sec:exp}
We evaluate GOFM from two complementary perspectives.  
First, we assess \emph{decision-time solving} on road graphs, where GOFM combines a pretrained structural prior with task-specific decoding to tackle shortest path (SP) and four NP-hard routing problems: Graphic-TSP, TP-SOD, TP-DOD~\citep{Martin2022ConstrainedSP}, and CVRP.  
Second, we conduct a \emph{real-world case study} on an operational delivery instance from the Amazon Last Mile Challenges dataset~\citep{Merchn20222021AL} to test whether the learned prior transfers to out-of-distribution logistics settings.  
Evidence on \emph{representation reuse} beyond routing (community detection and influence maximization) and a discussion of limitations are deferred to Section~\ref{sec:dis}, where they are interpreted in the broader context of GOFM's generalizability and practical scope.

\subsection{Datasets and Problem Construction}
\label{sec:data}

\paragraph{Road-graph datasets.}
We evaluate on four sparse road graphs spanning controlled and real-world settings: a synthetic simulation instance ($N{=}20$), a mid-scale urban network from Chengdu--Longquanyi ($N{=}132$), and two larger OpenStreetMap-derived road networks from Berkeley ($N{=}893$) and Palo Alto ($N{=}1945$). In all cases, we use an undirected primal graph $G=(V,E,w)$ in which nodes represent road intersections or endpoints, edges represent feasible road segments, and edge weights encode road-network travel costs. For real networks, geographic coordinates are retained for visualization, but optimization is always performed using graph-based costs implied by the underlying road topology rather than Euclidean distance. E-Companion Section~\ref{ec:road_graph_construction} provides the full construction procedure, preprocessing pipeline, and graph statistics.

\paragraph{Problem instances and evaluation metric.}
As defined in Section~\ref{sec:preliminary}, we study distance-based graph optimization on real-world sparse road networks. For all routing tasks, movement is restricted to the road graph $G=(V,E,w)$, and the travel cost between any two nodes $i,j\in V$ is measured by the graph-induced shortest-path distance
\[
d_G(i,j)\;=\;\min_{p:i\leadsto j}\sum_{(u,v)\in p} w_{uv}.
\]
This intentionally departs from complete-graph Euclidean proxies: $d_G$ captures detours, bottlenecks, and connectivity constraints that are intrinsic to real urban routing. All reported routing objectives are evaluated under this common network metric, rather than Euclidean complete-graph proxies.

Given a road graph $G$, we instantiate five routing tasks by sampling task specifications and optimizing the corresponding objective under $d_G$:
\begin{itemize}
    \item[] (1) Shortest Path (SP).
    An instance is defined by an origin--destination pair $(s,t)\in V\times V$, and the objective is to return a minimum-length $s{\to}t$ path.

    \item[] (2) Graphic Traveling Salesman Problem (Graphic-TSP).
    An instance is defined by a required node set $R\subseteq V$. The objective is a closed walk on $G$ that visits every node in $R$ at least once and minimizes total traversed edge length. Unlike metric TSP on complete graphs, vertex and edge repetitions may be necessary on sparse road networks; see \citet{Seb2012ShorterTB}.

    \item[] (3) Tour Problems with Required POIs (TP-SOD / TP-DOD).
    An instance is defined by endpoints $o,d\in V$ and a required POI set $R\subseteq V$. The objective is a minimum-length walk on $G$ that starts at $o$, visits all nodes in $R$ at least once, and ends at $d$. TP-SOD is the special case $o=d$; TP-DOD corresponds to $o\neq d$.
    This family is related to constrained tour-planning and tourist-trip problems
    \citep{Gunawan2016OrienteeringPA,Vansteenwegen2011TheOP,Martin2022ConstrainedSP}.

    \item[] (4) Capacitated Vehicle Routing Problem (CVRP).
    An instance is defined by a depot $o\in V$, a customer set $R\subseteq V$ with demands $\{q_u\}_{u\in R}$, vehicle capacity $Q$, and vehicle budget $m$. The objective is to construct up to $m$ depot-rooted routes that serve each customer exactly once, satisfy route capacities, and minimize total travel length under $d_G$.
\end{itemize}

Beyond these routing tasks, we also evaluate representation reuse on two non-routing graph problems, community detection and influence maximization, reported separately in Section~\ref{sec:repr_reuse_main}.

\subsection{Experimental Settings}
\label{sec:exp_settings}

\paragraph{Baselines.}
We compare against a broad set of baselines spanning exact solvers, classical heuristics, and modern learning-based approaches. For shortest path (SP), we include the label-setting algorithm of \citet{Dijkstra1959ANO} as the ground-truth optimum, together with A* heuristic search and Google's OR-Tools implementation~\citep{ortools}. For tour-based problems, we employ the state-of-the-art Lin--Kernighan--Helsgaun framework (LKH3)~\citep{Helsgaun2017AnEO}, the Christofides approximation followed by 2OPT local refinement, and simple greedy strategies such as Nearest Neighbor (NN). We also report Gurobi as a commercial MILP solver under fixed time limits, serving as a near-exact reference for NP-hard instances. On the learning side, we benchmark the Attention Model (AM) of \citet{Kool2018AttentionLT}, an RL-trained neural combinatorial optimization architecture under both greedy and sampling-based decoding, and the Instance-Conditioned Adaptation framework (ICAM)~\citep{Zhou2024InstanceConditionedAF}, which adapts neural solvers to different instance scales via lightweight conditioning modules. We further evaluate OR-LLM-Agent~\citep{Zhang2025ORLLMAgentAM}, which uses structured prompting and tool use, and two large language models, ChatGPT5 and Qwen3-235B~\citep{Yang2025Qwen3TR}, both in zero-shot settings with JSON-constrained outputs. For CVRP, we additionally include the Hybrid Genetic Search (HGS) algorithm~\citep{Vidal2020HybridGS} as a strong state-of-the-art VRP heuristic baseline.
Implementation details for road-graph preprocessing, visualization, and baseline adaptation to sparse road networks are deferred to E-Companion (EC) Sections~\ref{ec:data_processing}--\ref{ec:baseline_adaptation}.

\paragraph{Evaluation metrics.}
We report three metrics across all routing tasks: (\textbf{Obj}) objective value (total path/tour length; lower is better), (\textbf{Succ}) success rate (fraction of feasible outputs; higher is better), and (\textbf{Time}) average runtime per instance in seconds (lower is better).

\subsection{Main Results}
\label{sec:main_results}

\setlength{\tabcolsep}{2pt}
\begin{table*}[!htbp]
   \centering
   \caption{
   Performance across routing problems on Simulation ($N=20$),
   Chengdu ($N=132$), Berkeley ($N=893$), and Palo Alto ($N=1945$).
   Entries report objective, success (\%), and time (s).
   }
   \label{tab:main_table}
   \scriptsize

   \renewcommand{\arraystretch}{1.25}

   \resizebox{\textwidth}{!}{%
   \begin{tabular}{@{}c l ccc ccc ccc ccc@{}}
   \toprule \toprule
   & & \multicolumn{3}{c}{Simulation ($N=20$)}
     & \multicolumn{3}{c}{Chengdu ($N=132$)}
     & \multicolumn{3}{c}{Berkeley ($N=893$)}
     & \multicolumn{3}{c}{Palo Alto ($N=1945$)} \\
   \cmidrule(lr){3-5} \cmidrule(lr){6-8} \cmidrule(lr){9-11} \cmidrule(lr){12-14}
   & Method
     & Obj & Succ (\%) & Time (s)
     & Obj & Succ (\%) & Time (s)
     & Obj & Succ (\%) & Time (s)
     & Obj & Succ (\%) & Time (s) \\
   \midrule

   \multirow{7}{*}{\rotatebox{90}{SP}}
    & Dijkstra      & 6.09 & 100\% & 0.000 & 5.09 & 100\% & 0.000 & 2.23 & 100\% & 0.002 & 3.37 & 100\% & 0.003 \\
    & A*            & 6.09 & 100\% & 0.000 & 5.09 & 100\% & 0.000 & 2.23 & 100\% & 0.002 & 3.37 & 100\% & 0.005 \\
    & OR-Tools      & 6.09 & 100\% & 0.000 & 5.09 & 100\% & 0.000 & 2.23 & 100\% & 0.005 & 3.37 & 100\% & 0.012 \\
    \addlinespace[2pt]
    \cline{2-14}
    \addlinespace[2pt]
    & OR-LLM-Agent  & 15.22 & 30\% & 19.800 & -- & 0\% & -- & -- & 0\% & -- & -- & 0\% & -- \\
    & ChatGPT5      & 8.23  & 99\% & 3.394  & 8.85 & 62\% & 3.720  & -- & 0\% & --     & -- & 0\% & -- \\
    & Qwen3-235B    & 9.81  & 97\% & 1.890  & 11.98 & 45\% & 3.050  & -- & 0\% & --  & -- & 0\% & -- \\
    \cmidrule(lr){2-14} 
    & GOFM         & 6.24 & 99\% & 0.046
                    & 5.19 & 100\% & 0.821
                    & 2.43 & 98\% & 23.109
                    & 3.70 & 72\% & 59.61 \\
   \midrule\midrule

   \multirow{13}{*}{\rotatebox{90}{Graphic-TSP}}
    & LKH3          & 39.48 & 100\% & 0.016 & 87.50 & 100\% & 2.020 & 84.75 & 100\% & 2328.000 & 249.91 & 100\% & 20058.000 \\
    & Gurobi(30min) & 39.48 & 100\% & 0.147 & 90.64 & 100\% & 46.423 & 120.00 & 100\% & 1800.000 & -- & -- & -- \\
    & OR-Tools      & 39.48 & 100\% & 15.000 & 88.12 & 100\% & 30.000 & 109.78 & 100\% & 690.000 & 330.88 & 100\% & 10668.000 \\
    & Chr.\ + 2OPT  & 42.50 & 100\% & 0.007 & 97.78 & 100\% & 20.030 & 92.86 & 100\% & 606.502 & 267.33 & 100\% & 11658.800 \\
    \addlinespace[2pt]
    \cline{2-14}
    \addlinespace[2pt]
    & Greedy        & 44.43 & 100\% & 0.001 & 117.30 & 100\% & 0.006 & 113.76 & 100\% & 0.213 & 323.05 & 100\% & 0.463 \\
    & OR-LLM-Agent  & -- & 0\% & -- & -- & 0\% & -- & -- & 0\% & -- & -- & 0\% & -- \\
    & AM (greedy)   & 100.91 & 100\% & 0.042 & 104.53 & 100\% & 0.020 & -- & -- & -- & -- & -- & -- \\
    & AM (10)       & 82.16 & 100\% & 0.054 & 104.31 & 100\% & 0.017 & -- & -- & -- & -- & -- & -- \\
    & AM (100)      & 74.83 & 100\% & 0.160 & 101.35 & 100\% & 0.017 & -- & -- & -- & -- & -- & -- \\
    & AM (1000)     & 70.83 & 100\% & 1.366 & 100.36 & 100\% & 0.018 & -- & -- & -- & -- & -- & -- \\
    & ICAM          & 41.14 & 100\% & 0.159 & 103.25 & 100\% & 1.982 & 122.51 & 100\% & 571.172 & -- & -- & -- \\
    & Qwen3-235B    & 103.95 & 41\% & 2.690 & -- & 0\% & -- & -- & 0\% & -- & -- & 0\% & -- \\
    \cmidrule(lr){2-14} 
    & GOFM         & 39.48 & 100\% & 0.381
                    & 99.00 & 100\% & 6.405
                    & 99.26 & 100\% & 13.29
                    & 284.42 & 100\% & 77.75 \\
   \midrule\midrule

   \multirow{4}{*}{\rotatebox{90}{TP-SOD}}
    & Gurobi(30s)  & 23.96 & 100\% & 0.007 & 37.67 & 100\% & 0.159 & 25.79 & 100\% & 35.434 & 84.84 & 100\% & 180.192 \\
    & OR-Tools     & 23.96 & 100\% & 10.000 & 37.67 & 100\% & 15.058 & 25.73 & 100\% & 33.630 & 74.63 & 100\% & 114.418 \\
    & LKH3         & 23.96 & 100\% & 0.003 & 37.67 & 100\% & 0.137 & 25.63 & 100\% & 7.958  & 73.26 & 100\% & 119.820 \\
    \cmidrule(lr){2-14} 
    & GOFM        & 24.72 & 100\% & 0.030
                   & 39.62 & 100\% & 0.202
                   & 27.88 & 100\% & 3.352
                   & 80.19 & 100\% & 37.498 \\
   \midrule\midrule

   \multirow{4}{*}{\rotatebox{90}{TP-DOD}}
    & Gurobi(30s)  & 22.01 & 100\% & 0.004 & 34.46 & 100\% & 0.035 & 24.29 & 100\% & 11.672 & 73.23 & 100\% & 151.482 \\
    & OR-Tools     & 22.01 & 100\% & 10.003 & 34.73 & 100\% & 15.059 & 25.55 & 100\% & 34.666 & 73.37 & 100\% & 115.242 \\
    & LKH3         & 22.01 & 100\% & 0.002 & 34.04 & 100\% & 0.090 & 24.70 & 100\% & 10.751 & 69.44 & 100\% & 99.823 \\
    \cmidrule(lr){2-14} 
    & GOFM        & 22.87 & 100\% & 0.037
                   & 36.39 & 100\% & 0.083
                   & 27.76 & 100\% & 3.582
                   & 79.57 & 100\% & 35.030 \\
   \midrule\midrule

   \multirow{6}{*}{\rotatebox{90}{CVRP}}
    & OR-Tools            & 39.92 & 100\% & 10.000 & 57.66 & 100\% & 30.052 & 44.88 & 100\% & 32.750 & 167.05 & 100\% & 88.52 \\
    & LKH3                & 43.27 & 100\% & 1.589  & 59.61 & 100\% & 35.978 & 45.55 & 100\% & 257.649 & 168.55 & 100\% & 307.65 \\
    & Gurobi              & 39.92 & 100\% & 2.723  & 58.07 & 100\% & 46.610 & 47.48 & 100\% & 285.078 & 217.92 & 100\% & 320.59 \\
    & HGS(300s)           & 39.92 & 100\% & 300.01 & 57.65 & 100\% & 300.01 & 44.22 & 100\% & 300.01  & 162.25 & 100\% & 300.02 \\
    \cmidrule(lr){2-14}
    & GOFM(FI)    & 40.49 & 100\% & 0.290  & 60.35 & 100\% & 7.210  & 51.66 & 100\% & 4.630   & 228.74 & 100\% & 59.01 \\
    & GOFM(Split) & 42.62 & 100\% & 0.100  & 60.67 & 100\% & 1.340  & 47.50 & 100\% & 3.920   & 176.78 & 100\% & 34.42 \\
   \bottomrule \bottomrule
   \end{tabular}%
   }

   \par\vspace{0.8em}

   \begin{minipage}{0.95\textwidth}
   \footnotesize \textit{Note.}
   ``--'' has two meanings: when the success rate is $0\%$, the method produced outputs but none satisfied task constraints;
   when all three entries are ``--'', the method could not generate solutions for that problem setting at all.
   For Gurobi on the $N=1945$ instance, the solver ran out of memory and returned no feasible solution.
   \end{minipage}
   \vspace{-6pt}
\end{table*}

\begin{figure*}[t]
    \centering
    \begin{minipage}[t]{0.45\textwidth}
        \centering
        \includegraphics[width=\textwidth]{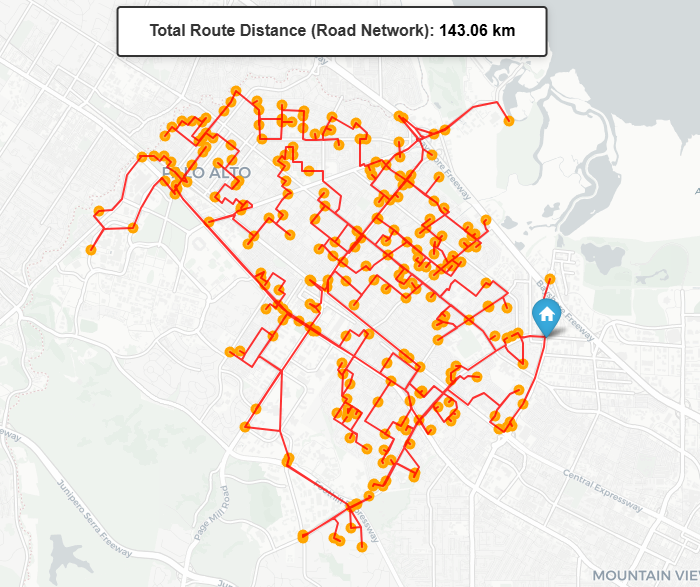}
        \vspace{2pt}
        {\scriptsize\textit{Panel A: Gurobi.}}
    \end{minipage}\hfill
    \begin{minipage}[t]{0.45\textwidth}
        \centering
        \includegraphics[width=\textwidth]{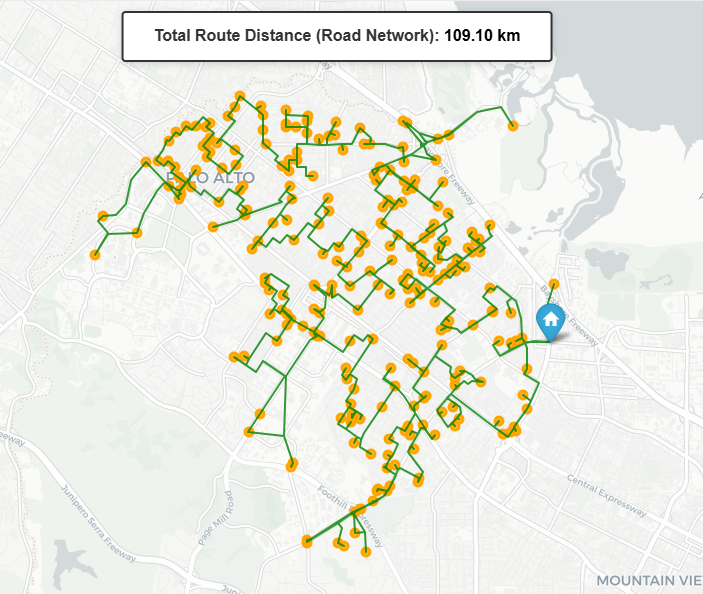}
        \vspace{2pt}
        {\scriptsize\textit{Panel B: GOFM.}}
    \end{minipage}
    \caption{A side-by-side comparison of TP-SOD solutions on the Palo Alto road network.}
    \label{fig:paloalto-example}
    \vspace{-10pt}
\end{figure*}

Table~\ref{tab:main_table} reports results across the simulation, Chengdu, Berkeley, and Palo Alto networks. We compare GOFM against three groups: classical solvers, neural baselines, and LLM-based methods.

\paragraph{GOFM vs.\ classical solvers: competitive quality, superior scaling.}
Across all five routing tasks, GOFM achieves near-optimal objective values with near-perfect feasibility. The efficiency advantage becomes most visible as graph size grows. On Graphic-TSP, solver-based runtimes already range from 606 to 2{,}328 seconds at the Berkeley scale ($N{=}893$). On the city-scale Palo Alto network ($N{=}1{,}945$), they rise to 10{,}668--20{,}058 seconds, whereas GOFM completes in 77.75 seconds, roughly two orders of magnitude faster. This speed does not come at the expense of solution quality: GOFM maintains $100\%$ feasibility for Graphic-TSP on both Berkeley and Palo Alto, with tour lengths that remain within a comparable range of the solver references (e.g., 99.26\,km vs.\ 84.75--120\,km on Berkeley). The same pattern extends to CVRP, where GOFM preserves $100\%$ feasibility across all networks with runtimes that stay practically small relative to solver budgets. The two CVRP decoders (FI and Split) expose a clear operational trade-off: Split is consistently faster, while FI tends to be more conservative in route construction, yet both preserve feasibility and competitive cost.

Why does GOFM scale so much better? The answer traces directly to the pretrain-and-decode architecture. Solver-based methods rely on iterative improvement and extensive neighborhood search, a strategy that becomes increasingly expensive on sparse, topologically constrained road graphs where local moves are often infeasible. GOFM, by contrast, generates solutions through its learned structural prior and task-specific decoding, pushing the expensive structural learning into a one-time pretraining cost. This separation of ``learning the graph'' from ``solving on the graph'' is the core mechanism behind the favorable scaling.

Figure~\ref{fig:paloalto-example} provides a qualitative look at a TP-SOD instance on the Palo Alto road network with $254$ mandatory visits. In this instance, GOFM achieves a $23.74\%$ reduction in route length relative to Gurobi (109.10\,km vs.\ 143.06\,km), while also reducing runtime from 320.35\,s to 90.76\,s. The comparison illustrates that GOFM can directly produce a coherent, globally structured tour from the learned prior, rather than relying on solver-style iterative local search that may get stuck in poor local optima on constrained sparse networks.

\paragraph{GOFM vs.\ neural baselines: sparse-graph robustness.}
Existing Transformer-based solvers such as the Attention Model~\citep{Kool2018AttentionLT} and ICAM~\citep{Zhou2024InstanceConditionedAF} are trained on task-specific synthetic distributions over complete Euclidean graphs. When transferred to sparse road networks, where edge structure, degree patterns, and the graph-induced metric all diverge sharply from training conditions, their solution quality drops substantially. On the simulation graph, AM produces Graphic-TSP tours roughly twice the length of the GOFM solution. ICAM does somewhat better but still falls well short on the Chengdu and Berkeley networks, and neither method scales to Palo Alto. This pattern confirms a structural limitation of instance-to-solution neural solvers: policies learned on complete-graph distributions lack the topological knowledge needed for feasible, high-quality decisions on sparse infrastructure graphs. GOFM avoids this problem entirely by learning the structural prior \emph{on the target graph itself}, making the model inherently aligned with the network's connectivity and distance structure.

\paragraph{GOFM vs.\ LLM-based methods: feasibility at scale.}
The LLM-based baselines (ChatGPT5, Qwen3-235B, OR-LLM-Agent) reveal a striking pattern: their success rate collapses as graph size increases. On SP, ChatGPT5 drops from $99\%$ feasibility at $N{=}20$ to $62\%$ at $N{=}132$ and $0\%$ at $N{=}893$ and $N{=}1945$. OR-LLM-Agent fails to produce any feasible solution on all but the smallest instances. This brittleness is not a coincidence. It reflects a basic representation mismatch. Language models operate over token sequences that carry statistical flexibility but have no built-in notion of graph adjacency, edge weights, or path connectivity. As the constraint space grows more complex, unconstrained text generation becomes increasingly unlikely to satisfy hard structural requirements. GOFM addresses this mismatch at the architectural level: because the model is pretrained directly on graph-consistent trajectories and paired with explicit constraint enforcement during decoding, feasibility is maintained by design rather than by luck. The contrast points to a broader lesson for the emerging LLM-for-optimization literature: on problems governed by strict structural constraints, graph-native representations are not just preferable but necessary for reliable performance at scale.

\subsection{Ablation Studies}
\label{sec:ablation}

The main results show that GOFM delivers competitive quality with favorable scaling. A natural follow-up question is: where does this performance come from? We run two targeted ablations to isolate the roles of the pretrained prior (encoder) and the decoder.

\paragraph{Ablation I: Is the gain from pretraining? (Random-walk prior vs.\ pretrained prior).}
We test whether the pretrained prior is essential by replacing it with a non-learned heuristic baseline. For each instance, we run $M$ distance-biased random walks on the same graph and compute empirical node visit frequencies, which are then used as proposal probabilities inside the same decoding routine. This random-walk (RW) prior is context-free and relies purely on local walk statistics.

\begin{table}[t]
\centering
\caption{Ablation on the value of pretraining for Graphic-TSP on the simulation graph ($N{=}20$).}
\label{tab:ablation_rw_prior}
\scriptsize
\setlength{\tabcolsep}{6pt}
\begin{tabular}{lccc}
\toprule
Method & Obj (km) & Gap (\%) & Time (s) \\
\midrule
Biased-RW Prior ($M{=}10$)   & 66.16 & 67.58 & 0.010 \\
Biased-RW Prior ($M{=}50$)   & 46.50 & 17.78 & 0.050 \\
Biased-RW Prior ($M{=}100$)  & 40.05 & 1.44  & 0.100 \\
GOFM(pretrained prior)      & \textbf{39.48} & \textbf{0.00} & \textbf{0.046 }\\
\bottomrule
\end{tabular}

\vspace{0.4em}
\begin{minipage}{0.96\linewidth}
\scriptsize
\textit{Note.} The decoding routine is kept identical; only the proposal prior is changed.
The Biased-RW prior is computed from $M$ distance-biased random walks and uses empirical visit frequencies
as proposal probabilities inside the same decoder.
Gap (\%) is computed relative to the best objective value in this table (39.48 km).
\end{minipage}
\vspace{-8pt}
\end{table}

Table~\ref{tab:ablation_rw_prior} reveals a clear pattern: increasing $M$ steadily strengthens the RW prior, but even at $M{=}100$ it still trails the pretrained GOFM prior under a comparable runtime budget (40.05\,km vs.\ 39.48\,km, with the RW baseline requiring more than twice the wall-clock time). The takeaway is that pretraining distills a much richer structural signal than the RW prior. In particular, it captures global connectivity, subpath compatibility, and metric geometry more effectively than raw visit-frequency statistics recovered from a finite sample of random walks.

A practical caveat reinforces this point: the RW ablation is only feasible on very small graphs. On realistic road networks with hundreds to thousands of nodes, biased random walks have very low probability of covering the required nodes under practical sampling budgets, and the computation needed to reach useful coverage grows quickly with graph size. The pretrained GOFM prior, by contrast, compresses the structural information of the entire graph into a learned parametric model, making it effective at scales where the RW heuristic becomes impractical.

\paragraph{Ablation II: Decoding strategy (CVRP).}
To isolate the contribution of the decoding algorithm, we keep the pretrained encoder fixed and vary only the decoder. We compare our structured GOFM decoding pipelines against a purely greedy decoder that deterministically selects the stepwise $\arg\max$ action without structured construction or local improvement. With the encoder fixed across conditions, any performance gap reflects the decoder’s value.

\begin{table}[t]
\centering
\caption{Ablation on decoding strategy for CVRP with the same pretrained encoder (structural prior). We report objective value, relative gap to the best objective in each column, and runtime.}
\label{tab:ablation_decoder_vrp}
\scriptsize
\setlength{\tabcolsep}{2pt}
\renewcommand{\arraystretch}{0.92}
\begin{tabular}{l ccc ccc ccc ccc}
\toprule
& \multicolumn{3}{c}{$N{=}20$} 
& \multicolumn{3}{c}{$N{=}132$}
& \multicolumn{3}{c}{$N{=}893$}
& \multicolumn{3}{c}{$N{=}1945$} \\
\cmidrule(lr){2-4}\cmidrule(lr){5-7}\cmidrule(lr){8-10}\cmidrule(lr){11-13}
Decoding 
& Obj & Gap (\%) & Time (s)
& Obj & Gap (\%) & Time (s)
& Obj & Gap (\%) & Time (s)
& Obj & Gap (\%) & Time (s) \\
\midrule
Greedy
& 46.74  & +15.4 & 0.02
& 91.50  & +51.6 & 0.09
& 103.78 & +118.5 & 0.58
& 318.39 & +80.1 & 1.56 \\
FI
& \textbf{40.49}  & \textbf{+0.0} & \textbf{0.290}
& \textbf{60.35}  & \textbf{+0.0} & \textbf{7.210}
& 51.66  & +8.8 & 4.630
& 228.74 & +29.4 & 59.01 \\
Split
& 42.62  & +5.3 & 0.100
& 60.67  & +0.5 & 1.340
& \textbf{47.50}  & \textbf{+0.0} & \textbf{3.920}
& \textbf{176.78} & \textbf{+0.0} & \textbf{34.42} \\
\bottomrule
\end{tabular}

\vspace{0.4em}
\begin{minipage}{0.96\linewidth}
\scriptsize
\textit{Note.} All rows share the same pretrained encoder; only the decoder differs.
Gap (\%) is computed relative to the best (lowest) objective within each $N$ column.
\end{minipage}
\vspace{-6pt}
\end{table}

Table~\ref{tab:ablation_decoder_vrp} confirms that structured decoding is essential for high-quality CVRP solutions on sparse road graphs. The greedy decoder is consistently and substantially worse, and the degradation accelerates with graph scale: the gap widens from $+15.4\%$ at $N{=}20$ to $+51.6\%$ ($N{=}132$), $+118.5\%$ ($N{=}893$), and $+80.1\%$ ($N{=}1{,}945$). The reason is intuitive: stepwise $\arg\max$ is myopic. It cannot enforce route-level structure (capacity-aware partitioning, global ordering) and cannot recover from early mistakes once the partial solution has solidified. On sparse graphs, where feasible alternatives are already limited by topology, the cost of myopic errors compounds quickly. Our structured decoders address this by imposing global construction bias (farthest-insertion or giant-tour-plus-split) and then applying local improvement to repair early decisions.

Taken together, the two ablations offer a clear decomposition of GOFM's performance. The pretrained prior contributes global structural knowledge that raw heuristics cannot match (Ablation~I), and the structured decoder translates that knowledge into feasible, high-quality solutions by enforcing route-level coherence (Ablation~II). Neither component alone is sufficient. Competitive performance requires their combination.

\subsection{Real-World Case Study: Amazon Last Mile Routing}
\label{sec:amazon_case}

To assess whether GOFM's learned prior transfers beyond controlled benchmarks, we conduct a case study on an operational delivery instance from the Amazon Last Mile Challenges dataset~\citep{Merchn20222021AL}, which contains historical routes executed by human drivers under real-world constraints.

We focus on a single instance from the Boston DBO2 station, evaluating the actual delivery sequence and the GOFM-generated solution under identical conditions:
(i) the same depot,
(ii) the same set of customer stops,
(iii) the same number of routes (vehicles),
and (iv) the same road-network distance metric.
Historical delivery sequences may reflect objectives beyond distance minimization, including driver experience, service policies, and time-dependent operational factors. Results should be interpreted accordingly.

\paragraph{Road network construction and training data separation.}
A key design point is that GOFM is not trained on any road-network artifacts released with the Amazon challenge. Instead, we reconstruct the street network directly from OpenStreetMap using \texttt{OSMnx}, extracting a bounding box that covers the service region and working with the resulting directed road graph. Depot and stop GPS coordinates are mapped to nearest road-network nodes via \texttt{cKDTree} queries, and the graph is augmented by inserting a depot node connected to multiple nearby road nodes to avoid a single-attachment bottleneck. Both solutions are evaluated under the same induced network metric. The results thus reflect an out-of-distribution transfer test: GOFM learns structural regularities from the road topology itself, rather than exploiting any dataset-specific preprocessing pipeline.


\begin{table}[t]
\centering
\caption{Route-level comparison on a real-world Amazon Last Mile instance (Boston DBO2). Distances are computed using road-network shortest paths (lower is better).}
\label{tab:amazon_route_level}
\scriptsize
\setlength{\tabcolsep}{8pt}
\renewcommand{\arraystretch}{0.92}
\begin{tabular}{l cc cc c}
\toprule
& \multicolumn{2}{c}{Actual Delivery Sequence}
& \multicolumn{2}{c}{GOFM Generated}
& \multirow{2}{*}{$\Delta$Dist. (\%)} \\
\cmidrule(lr){2-3}\cmidrule(lr){4-5}
Route
& \#Stops & Dist. (km)
& \#Stops & Dist. (km)
& \\
\midrule
Route 1 & 142 & 68.60 & 129 & 40.60 & $-40.8$ \\
Route 2 & 188 & 75.77 & 182 & 57.73 & $-23.8$ \\
Route 3 & 171 & 56.57 & 177 & 52.78 & $-6.7$  \\
Route 4 & 149 & 75.30 & 162 & 57.33 & $-23.9$ \\
\midrule
\textbf{Total}
& 650 & \textbf{276.24}
& 650 & \textbf{208.44}
& \textbf{$-24.5$} \\
\bottomrule
\end{tabular}

\vspace{0.4em}
\begin{minipage}{0.96\linewidth}
\scriptsize
\textit{Note.} ``Actual Delivery Sequence'' reports the observed route sequence in the Amazon instance, while ``GOFM Generated'' reports the corresponding GOFM solution. $\Delta$Dist. (\%) is the percentage distance change of GOFM relative to actual route.
\end{minipage}
\vspace{-15pt}
\end{table}

\begin{figure}[!htbp]
    \centering
    \begin{minipage}[t]{0.32\linewidth}
        \centering
        \includegraphics[width=\linewidth]{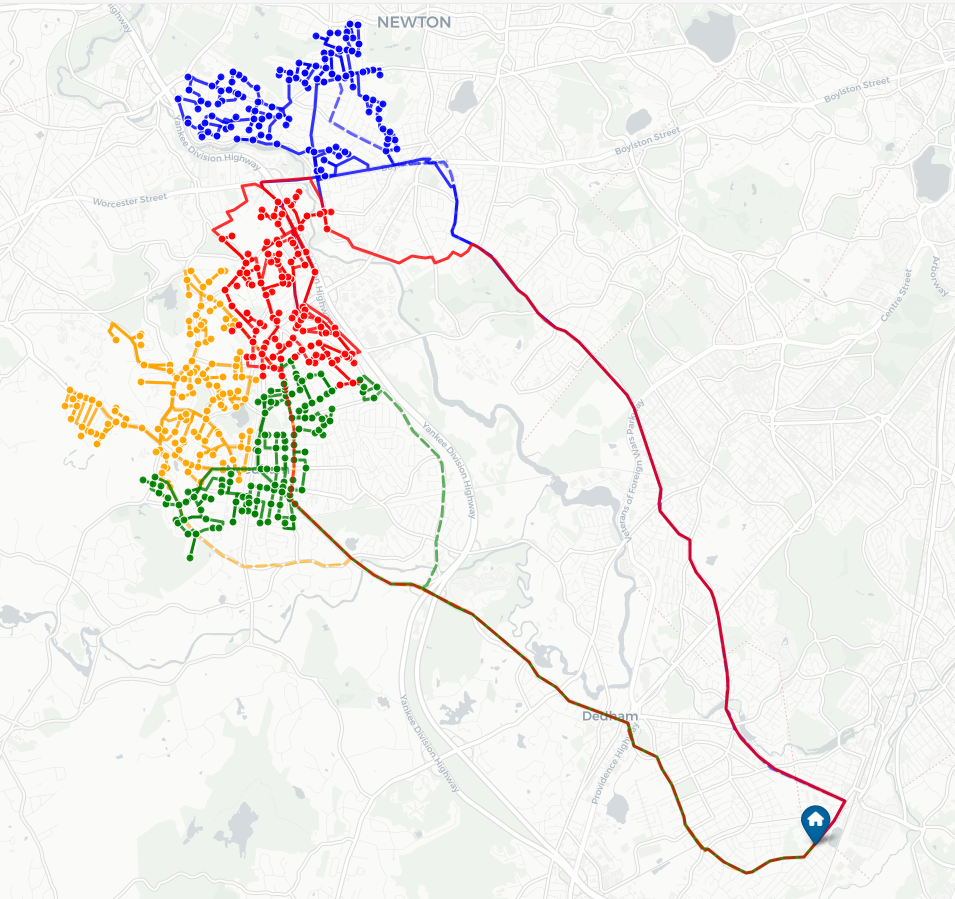}
        \vspace{2pt}
        {\scriptsize\textit{Actual Delivery Sequence.}}
    \end{minipage}\hfill
    \begin{minipage}[t]{0.32\linewidth}
        \centering
        \includegraphics[width=\linewidth]{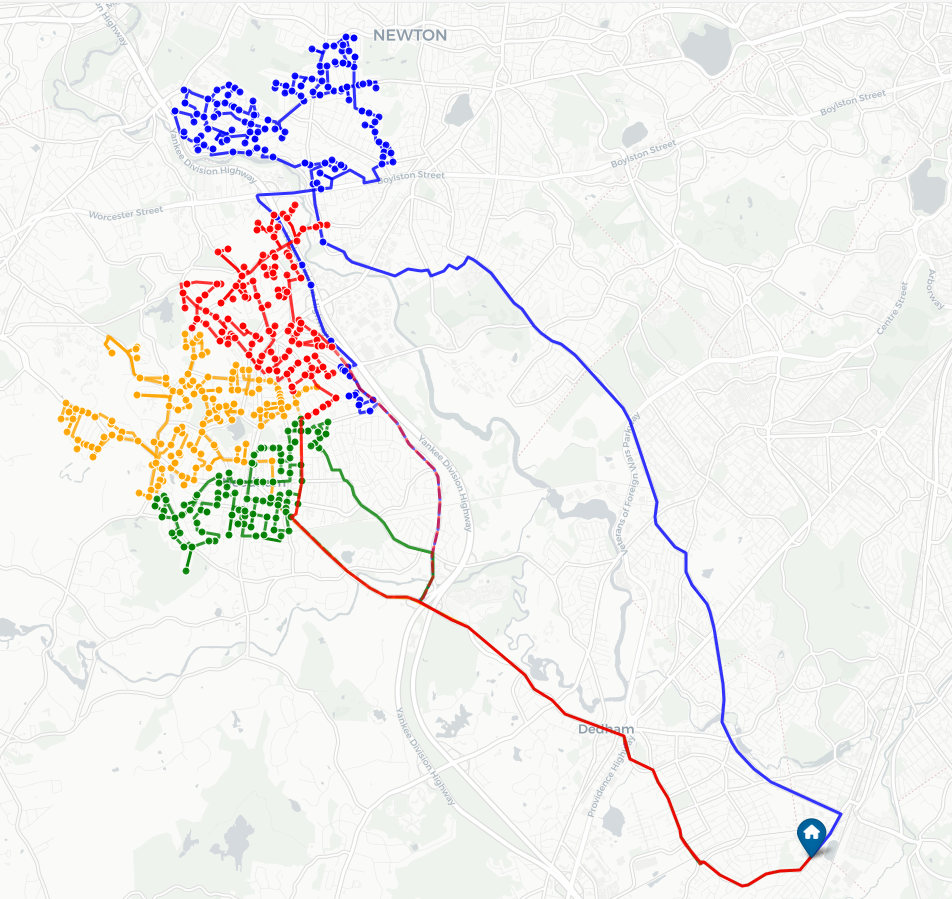}
        \vspace{2pt}
        {\scriptsize\textit{GOFM Generated.}}
    \end{minipage}\hfill
    \begin{minipage}[t]{0.32\linewidth}
        \centering
        \includegraphics[width=\linewidth]{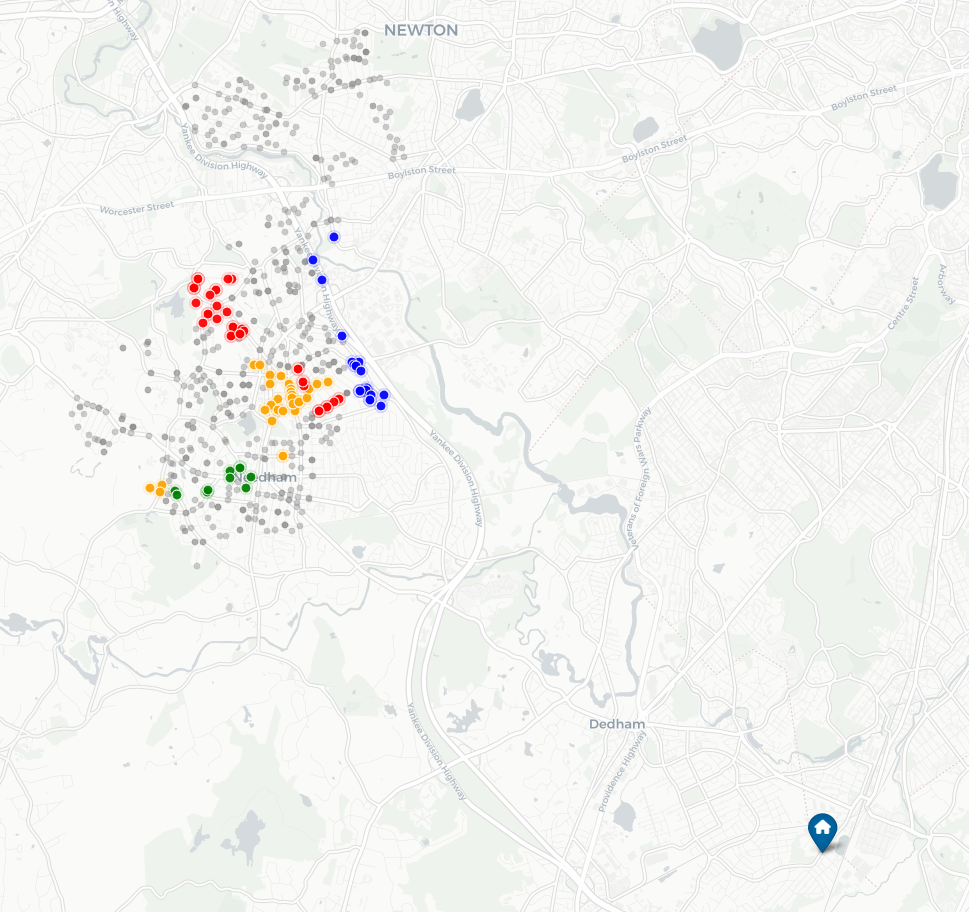}
        \vspace{2pt}
        {\scriptsize\textit{Difference view.}}
    \end{minipage}

    \caption{Amazon Last Mile routing visualization (Boston DBO2).}
    \label{fig:amazon_routes_three}

    \vspace{1pt}
    \begin{minipage}{0.98\linewidth}
        \scriptsize
        \textit{Note.} The historical sequence and the GOFM solution are evaluated under identical operational conditions (same depot, customer set, and fleet size).
        The difference view highlights divergences in stop-to-route assignment and route geometry.
    \end{minipage}
    \vspace{-15pt}
\end{figure}

\paragraph{Results and interpretation.}
As shown in Table~\ref{tab:amazon_route_level}, GOFM reduces total travel distance from 276.24\,km to 208.44\,km, a $24.5\%$ reduction. Distance improvements are observed consistently across all four routes (ranging from $-6.7\%$ to $-40.8\%$), indicating that the gain is systemic rather than driven by a single extreme reassignment.

Beyond distance reduction, the GOFM solution also reassigns stops across routes, suggesting that the learned structural prior supports route-set restructuring under a fixed fleet size. As Figure~\ref{fig:amazon_routes_three} shows, GOFM recovers route territories that closely resemble the actual service regions in the delivery plan, despite being trained solely on the OSM-derived road graph. The main differences arise from (i) stop-to-route reassignment near region boundaries and (ii) within-route visiting order. These are precisely the degrees of freedom that affect travel distance under a fixed depot, stop set, and fleet size. The case study thus highlights a practical capability: GOFM can reproduce operationally plausible route partitions while improving routing efficiency under the same constraints.

\paragraph{Back-of-the-envelope cost translation.}
To give the distance reduction an operational anchor, the per-instance saving is $\Delta D = 276.24 - 208.44 = 67.80$\,km. Using an illustrative operating-cost range of $c_{\mathrm{km}}\in[0.5,1.5]$\,USD/km (covering fuel/energy and distance-related vehicle costs, but excluding labor, service-time, and scheduling effects), the implied savings per route set are
\[
\widehat{\Delta C}_{\mathrm{instance}} = \Delta D \cdot c_{\mathrm{km}}
\in [33.9,\,101.7]\ \text{USD}.
\]
If a station executes $N_{\mathrm{ops}}$ comparable route sets per year, the annualized proxy is $\widehat{\Delta C}_{\mathrm{annual}} = \Delta D \cdot c_{\mathrm{km}} \cdot N_{\mathrm{ops}}$.
With $N_{\mathrm{ops}}=300$, this yields approximately \$10.2k--\$30.5k in annualized distance-related savings for this route-set pattern. We stress that this is not a full cost estimate: actual logistics costs depend on additional factors such as labor, time windows, congestion, parking, and fleet composition. The figure is offered as an order-of-magnitude indicator of operational relevance, not as a precise financial projection.

\section{Discussions}\label{sec:dis}

In this section, we step back from the numerical results to examine the generalizability of the learned prior beyond routing (Section~\ref{sec:repr_reuse_main}), the managerial and operational implications (Section~\ref{sec:managerial}), and the boundary conditions of the current GOFM design (Section~\ref{sec:limitations}).
\subsection{Generalizability Beyond Routing}
\label{sec:repr_reuse_main}

A foundation model should, by definition, provide value beyond the tasks it was designed for. Although GOFM focuses on distance-based routing and tour construction, we test whether the pretrained encoder produces node representations that are useful for structurally different graph analytics. The goal is not to claim state-of-the-art performance on these tasks, but to check whether the structural prior captures graph regularities that transfer to non-routing settings. Detailed protocols and implementation settings are provided in Section~\ref{ec:repr_reuse}.

\paragraph{Community detection (CD).}
We evaluate whether GOFM embeddings capture mid-level structure useful for clustering. On the American College Football network~\citep{Girvan2001CommunitySI}, we extract node embeddings from the pretrained encoder (Section~\ref{ec:repr_reuse}, Eq.~\eqref{eq:ec_node_embedding}) and run $K$-means with $K{=}12$ (matching the number of ground-truth conferences). Following prior work/survey on graph clustering and community detection~\citep{Zhou2022ACS,Park2022CGCCG}, we report Adjusted Rand Index (ARI) and Normalized Mutual Information (NMI). For both, higher values indicate better agreement with the ground-truth community labels.

\begin{table}[!htbp]
\vspace{-12pt}
\centering
\caption{Community detection results on the American College Football network.}
\label{tab:ec_community}
\scriptsize
\setlength{\tabcolsep}{5pt}
\begin{tabular}{lcc}
\toprule
Method & ARI & NMI \\
\midrule
Louvain (modularity maximization) & 0.9092 & 0.9764 \\
Label Propagation & 0.3686 & 0.7343 \\
Greedy Modularity & 0.5301 & 0.7624 \\
DeepWalk + k-means & 0.6923 & 0.8439 \\
node2vec + k-means & 0.9216 & 0.9545 \\
Adjacency embedding + k-means & 0.9206 & 0.9543 \\
\textbf{GOFM (ours)} & \textbf{0.9618} & \textbf{0.9774} \\
\bottomrule
\end{tabular}
\vspace{-12pt}
\end{table}

Table~\ref{tab:ec_community} shows that GOFM embeddings yield strong clustering quality. GOFM achieves the highest ARI among all methods and an NMI that matches or exceeds Louvain, despite the fact that community detection is not a training objective. Compared with standard walk-based embeddings (DeepWalk, Node2Vec), GOFM achieves a notably higher ARI (0.9618 vs.\ 0.9216), suggesting that the masked-trajectory reconstruction objective produces representations that are more sensitive to community boundaries than local co-occurrence statistics alone. This is intuitive: GOFM's progressive masking forces the model to reason about which graph regions can plausibly connect under the weighted metric, and that is exactly the kind of knowledge that distinguishes communities.

\paragraph{Influence maximization (IM).}
We further test whether the same embeddings support a simple heuristic for influence maximization. On the Berkeley road network, we select $k$ seeds using a heuristic built on GOFM node embeddings and evaluate expected spread under the Independent Cascade (IC) model via Monte Carlo simulation (details in Section~\ref{ec:repr_reuse}).

\begin{table}[!htbp]
\centering
\vspace{-12pt}
\caption{Influence maximization on the Berkeley road network (higher is better).}
\label{tab:ec_im}
\scriptsize
\setlength{\tabcolsep}{5pt}
\begin{tabular}{cccccc}
\toprule
$k$ & Random & Degree & PageRank & Greedy-IC & GOFM-IM \\
\midrule
0  & 0.00 & 0.00 & 0.00 & 0.00 & 0.00 \\
3  & 3.45 & 3.86 & 3.63 & 3.76 & 3.67 \\
6  & 7.05 & 7.73 & 7.19 & 7.52 & 7.11 \\
9  & 10.63 & 11.57 & 10.78 & 11.24 & 10.70 \\
12 & 14.16 & 15.21 & 14.34 & 14.86 & 14.13 \\
15 & 17.74 & 18.89 & 17.80 & 18.37 & 17.73 \\
18 & 21.39 & 22.35 & 21.20 & 21.99 & 21.37 \\
21 & 24.86 & 25.97 & 24.80 & 25.67 & 25.03 \\
\bottomrule
\end{tabular}
\vspace{-12pt}
\end{table}

Table~\ref{tab:ec_im} shows that the GOFM-based heuristic consistently outperforms random seeding and remains competitive with degree and PageRank heuristics across all budget levels, approaching the greedy IC reference. Together with the community detection results, these findings provide preliminary but encouraging evidence that the pretrained GOFM encoder can serve as a reusable graph representation backbone for tasks beyond routing. The structural prior, learned from trajectory reconstruction on a single graph, appears to encode regularities (centrality, community structure, connectivity bottlenecks) that are useful across quite different problem classes.

\subsection{Managerial and Operational Implications}
\label{sec:managerial}

The experimental evidence, from controlled benchmarks through the Amazon case study, points to a practical insight with direct managerial relevance: a persistent network can be treated as a \emph{reusable decision asset}. In many organizations, the same physical infrastructure graph supports a stream of related but different decisions (food delivery, parcel distribution, inspection tours, emergency response), yet these decisions are typically handled by separate tools, repeated solver tuning, and task-specific heuristic engineering. GOFM suggests an alternative: learn a graph-specific structural prior once, and reuse it across multiple optimization workflows with lightweight decoding.

This design has three concrete implications for operations management. First, it reduces repeated modeling effort. Rather than configuring a new solver pipeline for each routing variant, a single pretrained backbone supports SP, tour families, and CVRP through changes in the decoding layer alone. For logistics operators that regularly face new planning scenarios on the same service region (a common pattern in urban delivery, transit, and maintenance), this consolidation can translate into faster deployment and lower engineering overhead. Second, GOFM's inference speed (seconds to minutes vs.\ hours for solvers on city-scale graphs) makes it suitable for time-sensitive decisions where exact solvers are too slow: real-time dispatching, rolling-horizon re-optimization, and rapid what-if analysis under disruptions. Third, the representation-reuse results (Section~\ref{sec:repr_reuse_main}) suggest that the same pretrained model could support not only routing decisions but also network analytics such as community segmentation and influence-based targeting, creating a shared decision intelligence layer across planning and operational functions.

We note that these implications are strongest in settings where the underlying infrastructure evolves slowly relative to the decision queries it supports. Organizations operating across many distinct and rapidly changing networks may not yet benefit from the current design without extensions for cross-graph transfer (see Section~\ref{sec:limitations}).

\subsection{Limitations and Boundary Conditions}
\label{sec:limitations}

Our results should be interpreted within the scope of the current GOFM design. Three boundary conditions are particularly important.

First, GOFM is tailored to \emph{distance-based graph optimization} on real networks, where feasible decisions can be naturally expressed as paths, tours, or route sets and where a graph-conditioned structural prior can be learned from random-walk trajectories. The current traversal-based pretraining and decoding interface is well aligned with routing and tour-family problems but is less directly applicable to partitioning- or assignment-style combinatorial problems (such as Max-Cut, graph coloring, or matching) that lack a natural traversal formulation. Extending GOFM to these problem classes would require new pretraining and decoding schemes beyond trajectory reconstruction.

Second, the framework is designed for \emph{single-graph, multi-task reuse}, which matches many OR deployment settings on persistent infrastructure networks but also makes the learned prior graph-specific. In its current form, GOFM is not a universal cross-graph foundation model. Transferring to a new graph requires rebuilding the trajectory corpus and retraining (or at minimum re-adapting) the model. Whether part of the learned structural prior can generalize across related networks, for instance across road networks of similar urban type, is an open question and an important direction.

Third, our evaluation, while spanning four networks from $N{=}20$ to $N{=}1{,}945$ and covering five distinct routing tasks plus two graph analytics problems, is necessarily bounded. We have not yet tested GOFM on time-dependent or stochastic edge weights, on graphs with tens of thousands of nodes, or on problems with rich side constraints (e.g., time windows, mixed fleets, or pickup-and-delivery pairing). Each of these extensions would stress different aspects of the pretrain-and-decode architecture and may require methodological refinements beyond the current framework.

\section{Conclusions}\label{sec:conclusion}
We presented the Graph Optimization Foundation Model (GOFM), a graph-native pretrain--decode framework that adapts the foundation-model paradigm to operations research problems on graph networks. Rather than training a separate solver or neural policy for each downstream task, GOFM learns a reusable structural prior from self-supervised, structure-aware random-walk trajectories on a fixed graph and then combines this prior with lightweight, task-specific, constraint-aware decoding to produce feasible solutions. 

Three contributions emerge from this work. First, we showed that the pretrain-transfer logic underlying large language models can be applied to graph optimization when three design choices are made jointly: a graph-consistent trajectory corpus, a progressive masked-reconstruction pretraining objective, and explicit constraint-aware decoding. The ablation studies confirm that both the pretrained prior and the structured decoder are necessary, and that neither component alone accounts for GOFM's performance. Second, the experimental results show that GOFM achieves competitive solution quality relative to state-of-the-art solvers across five routing task families and four network scales, with inference times one to two orders of magnitude faster on city-scale graphs. Importantly, this speed does not come at the cost of feasibility: GOFM maintains near-perfect constraint satisfaction across all settings, a property that neither neural solvers trained on complete-graph distributions nor LLM-based pipelines can sustain as problem size grows. The Amazon Last Mile case study further shows that the learned prior transfers to an out-of-distribution operational setting, yielding a
24.5\% distance reduction without task-specific retraining. Third, the representation-reuse experiments on community detection and influence maximization provide preliminary evidence that the structural prior captures graph regularities useful beyond routing, reinforcing the ``foundation" identity of the framework.

At the application level, the pretrain--decode architecture is a natural fit for settings where a persistent network supports a recurring stream of different decisions under changing operating conditions. Urban logistics, transit planning, maintenance scheduling, and emergency response are all candidate domains. More broadly, our findings suggest that the value of graph-native pretraining lies not in replacing a single solver, but in building a shared structural-intelligence layer that reduces repeated modeling effort, speeds up decision cycles, and improves cross-task consistency across an organization's optimization workflows. We view GOFM not as a universal solver, but as a step toward reusable, graph-native decision intelligence for operations research and management practice.
Looking ahead, several research directions would extend the scope and impact of this paradigm. Incorporating time-dependent and stochastic edge weights would bring GOFM closer to dynamic operational environments where congestion, demand uncertainty, and real-time re-optimization matter most. Cross-graph transfer also remains an open and practically important question: can part of the learned prior generalize across structurally related networks without full retraining? This is especially relevant for organizations that manage families of similar infrastructure graphs, such as urban road networks across multiple cities.

\vspace*{.2in}

\bibliographystyle{plainnat}
\bibliography{reference}

\vspace*{.5in}
\noindent{\bf\Large Appendices: Additional Experiments and Implementation Details}
\appendix
\numberwithin{equation}{section}
\renewcommand{\theHsection}{EC.\arabic{section}}
\renewcommand{\theHsubsection}{EC.\arabic{section}.\arabic{subsection}}
\renewcommand{\theHsubsubsection}{EC.\arabic{section}.\arabic{subsection}.\arabic{subsubsection}}
\renewcommand{\theHequation}{EC.\arabic{equation}}
\renewcommand{\theHfigure}{EC.\arabic{figure}}
\renewcommand{\theHtable}{EC.\arabic{table}}
 


\section{Data Processing and Visualization}\label{ec:data_processing}

\subsection{Road-Graph Construction}\label{ec:road_graph_construction}

\paragraph{Datasets.}
Our evaluation uses one controlled synthetic graph and three real-world road networks to balance exact reproducibility with operational realism. Across all datasets, we model the environment as a sparse undirected primal graph $G=(V,E,w)$, where nodes represent road intersections or endpoints, edges represent valid road segments, and weights encode travel cost. For the real networks, each node retains its original WGS84 geographic coordinates for plotting and spatial sanity checks. Throughout, optimization costs are computed from edge lengths on the road network and therefore follow the graph-induced shortest-path metric rather than Euclidean distance.

\textbf{(1) Synthetic simulation graph ($N{=}20$).}
We construct a fixed sparse graph with $20$ nodes and $34$ edges (density $0.1789$, average degree $3.40$) to provide a controlled testbed in which topology and sparsity are exactly reproducible. To mimic heterogeneous traffic conditions, edge weights are sampled once under a fixed random seed ($s{=}42$): low-cost ``arterial'' edges are drawn from $U(1.0,3.0)$, whereas less preferred cross links are drawn from $U(3.5,7.0)$. The graph structure is held fixed, and node coordinates are used only for visualization.

\textbf{(2) Chengdu--Longquanyi ($N{=}132$).}
This self-collected dataset is a real road network from Longquanyi District, Chengdu, China. After preprocessing into an undirected primal graph, the instance contains $132$ nodes and $222$ edges (density $0.0257$, average degree $3.36$). Edge weights are obtained from original road-segment lengths in the source map data and converted from meters to kilometers for consistency with the other datasets. The resulting graph preserves realistic urban sparsity and nonuniform local connectivity.

\textbf{(3) Berkeley ($N{=}893$).}
We extract the Berkeley network from OpenStreetMap using \texttt{OSMnx}~v2.0+ with \texttt{network\_type=drive}. Specifically, we clip a $1.3$ km radius around the landmark ``Downtown Berkeley BART'' and retain only drivable streets. The raw OSM MultiDiGraph is then simplified into an undirected simple graph by removing self-loops, consolidating parallel edges by keeping the minimum length, preserving the largest connected component, and relabeling nodes to $\{0,\dots,N{-}1\}$. The final graph has $893$ nodes and $1{,}413$ edges (density $0.0035$, average degree $3.16$). Edge weights are road-segment geometric lengths reported in kilometers, and node coordinates $(\text{lon},\text{lat})$ are stored for visualization.

\textbf{(4) Palo Alto ($N{=}1945$).}
The Palo Alto graph is also obtained from OpenStreetMap via \texttt{OSMnx}~v2.0+ with \texttt{network\_type=drive}, but here we query the full place boundary rather than a radius around a landmark. We then apply the same standard preprocessing steps as above: simplification into an undirected simple graph, removal of self-loops, consolidation of parallel edges by minimum length, extraction of the largest connected component, and node relabeling to $\{0,\dots,N{-}1\}$. The resulting city-scale road graph contains $1945$ nodes and $2770$ edges. Edge weights are measured in kilometers from road-segment lengths, and GPS coordinates are retained for plotting.


\section{Additional Experimental Settings}
\label{ec:exp_settings}

\subsection{Hyperparameters and training details}
\label{ec:hyperparams}

\paragraph{Model backbone.}
Our distance-based Graph Optimization Foundation Model (GOFM) adopts a Transformer encoder with task-dependent depth/width and walk tokenization.
Across all settings we use Adam, dropout $=0.1$, and early stopping based on validation performance when applicable.

\subsection{Hardware, runtime measurement, and solver/LLM access}
\label{ec:runtime}

\paragraph{Compute environment.}
All experiments are run on an Intel i5-12400F CPU with 32GB RAM and an RTX 4060 Ti (8GB) GPU (CUDA 12.9).

\paragraph{Runtime measurement.}
We report wall-clock time measured on the same machine for all locally executed methods.
Classical solvers (Dijkstra, OR-Tools, LKH3, Gurobi) use their official implementations with fixed budgets/time limits as specified below.

\paragraph{Baseline configurations.}
Unless otherwise stated, we follow recommended default settings:
LKH3 with \texttt{MAX\_TRIALS}=10{,}000 and \texttt{RUNS}=10;
OR-Tools with \texttt{PATH\_CHEAPEST\_ARC} + \texttt{GUIDED\_LOCAL\_SEARCH};
Gurobi with MIP gaps in 0.1--5\% and time limits in 30--1800s;
A* with $c_{\text{puct}}=1.4$ and $\alpha=1.8$.
Additional adaptation details to sparse road networks are provided in Section~\ref{ec:baseline_adaptation}.

\paragraph{LLM baselines.}
ChatGPT5, OR-LLM-Agent, and Qwen3-235B are accessed via APIs.
We use zero-shot prompting with JSON-constrained outputs, and report end-to-end latency inclusive of API response time.

\section{Baseline Adaptation on Sparse Road Networks}\label{ec:baseline_adaptation}

Most classical solvers and some baselines (e.g., LKH3~\citep{Helsgaun2017AnEO}, OR-Tools, Gurobi, ICAM~\cite{Zhou2024InstanceConditionedAF})
assume a complete graph input, where every pair of nodes has a direct edge. However, real road networks are sparse:
many node pairs are not directly connected. To enable fair evaluation, we adopt two adaptation strategies:
metric closure and virtual edge construction.

\paragraph{Metric closure.}
Let $G=(V,E,w)$ be the original weighted road network. For any pair $u,v\in V$, define
\[
d(u,v) \;=\; \min_{p\in \mathcal{P}(u,v)} \sum_{(i,j)\in p} w(i,j),
\]
where $\mathcal{P}(u,v)$ is the set of all paths between $u$ and $v$.
The metric closure is the complete graph $G^\ast=(V,E^\ast,d)$ with edge weights equal to shortest-path distances in $G$.

\paragraph{Virtual edge construction.}
For path problems with distinct start and end nodes, e.g., TP-DOD, solvers like LKH3 require a closed TSP formulation.
We therefore introduce a virtual edge from the terminal node back to the start node with weight zero,
transforming the open-path problem into a tour. After solving, this artificial edge is removed and the remaining sequence
is expanded back into a feasible path in $G$.

\paragraph{Application.}
For both strategies, the adapted representation (closure graph $G^\ast$ or virtual edge construction) is fed to the baseline solver
(e.g., LKH3, OR-Tools, AM, ICAM) to generate a high-level tour $\pi^\ast$.
Each edge $(u,v)\in\pi^\ast$ is then expanded back into its shortest path in the original graph $G$,
guaranteeing feasibility while preserving the solver’s optimization logic.

\section{Decoding for graph optimization}
\label{ec:decoding}

At inference time, we translate the learned structural prior into task-feasible decisions through a decoding procedure that explicitly accounts for constraints. Formally, given a graph $G=(V,E,w)$ and a task specification $s$, we generate
\begin{equation}
\widehat{\mathbf{Y}} \;=\; \mathrm{Decode}\big(\pi(\cdot \mid G),\, s\big),
\end{equation}
where $\pi(\cdot\mid G)$ is the pretrained structural prior and $\mathrm{Decode}(\cdot)$ is a constructive procedure that produces a feasible structured solution (path/tour/walk). The key separation is that $\pi(\cdot\mid G)$ is task-agnostic. $\pi(\cdot\mid G)$ provides a task-independent structural prior tied only to $G$. Whereas feasibility and objective optimization for a given problem instance are enforced by $s$ during decoding. Thus, this design preserves transferability, once $\pi(\cdot\mid G)$ is learned, different optimization problems can be addressed by changing only the constraint logic and scoring used in decoding, rather than retraining the model.

\paragraph{Unified decoding template.}
We implement $\mathrm{Decode}(\cdot)$ as a constrained constructive procedure guided by the pretrained structural prior. At each step, the prior proposes promising next updates, while the task constraints and objective determine which proposal can be accepted.
Let $S_t$ be the current partial solution (e.g., a partial walk, partial tour, or partial ordering of required nodes).
Given $S_t$, we query the pretrained model to obtain a proposal distribution $P(\cdot \mid S_t, G)$ over candidate next-node updates. To keep decoding efficient and robust, we do not search over all nodes in $V$; instead, we form a small candidate set $\mathcal{C}_t$ (e.g., top-$K$ proposals under $P$ after basic admissibility filtering).

For each candidate $c\in\mathcal{C}_t$, we compute a task-specific incremental cost $\Delta J(c; S_t, G, s)$, which measures the objective change induced by applying update $c$ under the metric of $G$.
We then apply a feasibility operator $\Pi_s(\cdot)$ to enforce hard constraints specified by $s$.
The decoding step is
\[
S_{t+1}
=\Pi_s\!\Big(S_t \oplus \arg\min_{c\in\mathcal{C}_t}\Delta J(c;S_t,G,s)\Big),
\]
where $\oplus$ denotes the corresponding update operation (append, insert, or merge).
In this view, the structural prior narrows the search to a small set of high-quality directions, and the constraint/objective layer converts these directions into feasible, low-cost decisions. We next instantiate this template for specific problems on sparse road networks.

\paragraph{Shortest path decoding (SP).}
For SP with terminals $(o,d)$ on a sparse road network, the main challenge is to construct an $o{\to}d$ route that is globally coherent while avoiding pathological local behavior like short cycles that can arise from purely myopic expansions.
We therefore decode in a completion-style manner. We represent the current $o{\to}d$ walk as a sequence with unresolved interiors between anchor nodes, and recursively fill these masked gaps.
In this setting, the structural prior provides exactly the signal we need: it ranks which intermediate nodes are most compatible with the surrounding context.
So for each unresolved gap $g=(a,b)$ (between two consecutive anchors), we query the model to obtain $P(\cdot\mid \mathrm{ctx}(g),G)$ and restrict attention to a small candidate set
\[
\mathcal{C}(g) \;=\; \textsc{TopK}\!\Big(P(\cdot\mid \mathrm{ctx}(g),G)\Big).
\]
We then expand a beam of partial walks by inserting candidates from $\mathcal{C}(g)$, and to stabilize the expansion we employ a simple tabu-like local memory, an NG-set of recently visited nodes~\cite{Toriello2014ADT,Roberti2014DynamicNR}:
\[
\mathcal{C}_{\mathrm{NG}}(g) \;=\; \mathcal{C}(g)\setminus \mathrm{NG}(S_t),
\qquad
v_{t+1}\in \mathcal{C}_{\mathrm{NG}}(g),
\]
which filters repeated candidates and suppresses short cycles.
After a fixed expansion budget or when all gaps are resolved, we project each candidate walk to a feasible path on $G$. 
Formally, if a decoded walk contains a jump $(i,j)\notin E$, we replace it by the shortest path segment under the graph metric,
\[
(i,j)\;\mapsto\; \textsc{SP}_G(i,j), 
\qquad 
d_G(i,j)\;=\;\min_{p:i\leadsto j}\sum_{e\in p} w_e,
\]
and finally return the lowest-length feasible path among the beam, i.e.,
\[
\widehat{\mathbf{Y}} \;=\; \arg\min_{\mathbf{y}\in \mathcal{B}_{\mathrm{feas}}} \;\sum_{(i,j)\in \mathbf{y}} w_{ij}.
\]

\paragraph{Tour-family decoding (Graphic-TSP / TP-SOD / TP-DOD).}
For tour-based problems with a required node set $R$, decoding must address two coupled decisions at each iteration: (i) which unvisited node should be brought into the structure next, and (ii)where it should be inserted in the current partial tour/walk.

We adopt an insertion-style constructive heuristic because it naturally maintains a coherent partial structure throughout construction.
In particular, we follow a difficulty-first rationale common in routing heuristics: nodes that are peripheral or hard to accommodate are inserted early, since delaying them typically increases insertion regret once the partial structure has solidified, often forcing long detours or disruptive repairs.

Concretely, let the current partial tour be the anchor sequence $S_t=[a_1,\dots,a_L]$, with consecutive gaps $\mathcal{G}(S_t)=\{(a_\ell,a_{\ell+1})\}_{\ell=1}^{L-1}$, and $(a_L,a_1)$ for cycle variants.
For a remaining node $u\in R\setminus S_t$, we define a conservative compatibility (difficulty) score by aggregating the prior across all current gaps:
\[
\kappa_t(u)\;=\;\min_{g\in \mathcal{G}(S_t)} P\big(u \mid \mathrm{ctx}(g),G\big),
\qquad
u_t^\star \;=\; \arg\min_{u\in R\setminus S_t}\kappa_t(u).
\]
This selection inserts the most difficult-to-place node first, while the partial structure still has flexibility.
Given $u_t^\star$, we determine its insertion position by evaluating a small set of promising gaps suggested by the prior. Specifically, for each gap $g=(a,b)$ we obtain a compatibility score $P(u_t^\star\mid \mathrm{ctx}(g),G)$ and form a shortlist
\[
\mathcal{G}_t(u_t^\star)\;=\;\textsc{TopK}\!\Big(\{\,P(u_t^\star\mid \mathrm{ctx}(g),G)\,:\,g\in\mathcal{G}(S_t)\}\Big).
\]
Among these candidate gaps, we choose the position that minimizes the incremental travel length under the projected metric $d_G(\cdot,\cdot)$:
\[
g_t^\star \;=\; \arg\min_{(a,b)\in \mathcal{G}_t(u_t^\star)} 
\Delta J(u_t^\star;a,b),
\qquad
\Delta J(u;a,b)\;=\; d_G(a,u)+d_G(u,b)-d_G(a,b).
\]
We then insert $u_t^\star$ into $S_t$ at gap $g_t^\star$ and repeat until all required nodes are included.

Task variants enter only through constraint handling.
For TP-SOD, the tour must start and end at the same depot $o$, which is enforced by fixing $o$ as an immutable anchor throughout construction.
For TP-DOD with distinct $(o,d)$, we enforce an open walk by introducing a temporary virtual closure edge $(d,o)$ during construction, so the procedure operates on a cycle and reuses the same insertion logic, and then removing this virtual edge in the final output, yielding a feasible $o{\to}d$ walk that visits all required nodes.

\noindent\textbf{Why least-confidence-first?}
The rationale is to control downstream insertion regret.
A node that receives low compatibility under the current gaps is ``structurally hard'' for the partial tour: if postponed, it is likely to remain incompatible with most gaps after the tour solidifies, forcing long repairs or detours. By inserting the lowest-confidence node early (when the structure is still flexible), we preserve feasible low-cost placements
and reduce the chance of late-stage expensive insertions.

\paragraph{Vehicle routing (CVRP).}
For CVRP with depot $o$, customer set $R$ and demands $\{q_u\}_{u\in R}$ under vehicle capacity $Q$ and a vehicle budget $m$, capacity constraints make early choices particularly consequential. Following similar logic, inserting ``easy'' customers first can consume the remaining flexibility of the routes and leave a small set of difficult customers that are expensive or even infeasible to accommodate later.
We therefore instantiate the unified template using two practical decoders that differ only in the construction state and the feasibility projection, while leveraging the same pretrained structural prior for proposal generation.

\textit{(i) Global parallel insertion (multi-route construction).}
We maintain a partial route set $S_t=\{r_t^{(1)},\ldots,r_t^{(m)}\}$, where each route is an anchor sequence starting at $o$, and an unserved set $U_t=R\setminus \bigcup_k r_t^{(k)}$.
At each step, we query the prior on masked route gaps across all current routes to obtain next-node probabilities, which serve as compatibility signals between unserved customers and currently available insertion gaps.
To control downstream regret, we prioritize customers that are hard to place under the current partial solution.
Specifically, we aggregate compatibility conservatively across all queried gaps and select the customer whose best-case compatibility is weakest:
\[
\kappa_t(u)\;=\;\min_{g\in\cup_k \mathcal{G}(r_t^{(k)})} P\big(u \mid \mathrm{ctx}(g),G\big),
\qquad
u_t^\star \;=\;\arg\min_{u\in U_t}\kappa_t(u).
\]
This difficulty-first choice ensures that structurally uncertain customers are inserted while the route set still has flexibility, rather than being left to the end when only costly or infeasible options remain.
Given $u_t^\star$, we then select the route and position that minimize incremental travel length in the projected metric $d_G(\cdot,\cdot)$, subject to capacity feasibility.
For inserting $u$ between consecutive anchors $(a,b)$, the incremental travel cost is evaluated as
\[
\Delta J(u; a,b)\;=\; d_G(a,u)+d_G(u,b)-d_G(a,b),
\]
and $\Pi_s(\cdot)$ rejects insertions that violate $\sum_{v\in r_t^{(k)}} q_v + q_u \le Q$.
We repeat until $U_t=\emptyset$, yielding a capacity-feasible set of $m$ routes.

\textit{(ii) Giant-tour + split (capacity-aware projection).}
Alternatively, we decouple \emph{ordering} from \emph{capacity feasibility}.
We first decode a single customer ordering (a ``giant tour'') over $R$ using the same insertion-style tour-family decoder, leveraging the prior to obtain an ordering that is structurally compatible with the underlying road network.
We then project this permutation into a feasible CVRP solution via a standard split dynamic program that partitions the sequence into at most $m$ capacity-feasible routes~\citep{Vidal2020HybridGS}.
Each segment cost is computed using shortest-path distances in $G$, and any segment exceeding capacity is disallowed by $\Pi_s(\cdot)$.
Finally, we optionally apply lightweight local improvement in the same projected metric (e.g., 2-opt within routes and relocate/swap/2-opt$^\star$ across routes).

\paragraph{Remarks.}
Across tasks, decoding is polynomial-time and does not rely on exhaustive search.
The pretrained prior $\pi(\cdot\mid G)$ provides a reusable structural bias for proposing graph-consistent candidates, while feasibility is guaranteed (in practice) by explicit projection onto paths/walks in the underlying sparse road network.

\subsection{Representation Reuse Beyond Routing Tasks}
\label{ec:repr_reuse}

This appendix provides reproducible protocols for two classic graph mining analyses (community detection and influence maximization) based on the pretrained GOFM encoder.

\paragraph{Node embeddings from contextual walk states.}
Given a trained GOFM encoder $f_\theta$, we construct a node embedding for each node $i\in V$ by aggregating its contextual hidden states over biased random walks. Specifically, for each node $i$ we generate $M$ biased random walks of length $L$ starting from $i$:
\[
w_i^{(m)}=\big(v_0^{(m)}=i, v_1^{(m)},\ldots, v_{L-1}^{(m)}\big),\qquad m=1,\ldots,M.
\]
Each walk is fed into the encoder to obtain contextual hidden states
\[
H^{(m)} = f_\theta\!\left(w_i^{(m)}\right)\in\mathbb{R}^{L\times d}, \qquad H_t^{(m)}\in\mathbb{R}^d.
\]
We then collect all positions where the token equals node $i$:
\[
S_i=\{(m,t): v_t^{(m)} = i\},
\]
and define the node embedding as the average of its contextual hidden states:
\begin{equation}
e_i \;=\; \frac{1}{|S_i|}\sum_{(m,t)\in S_i} H_t^{(m)} \in \mathbb{R}^d.
\label{eq:ec_node_embedding}
\end{equation}

\paragraph{Community detection protocol.}
We evaluate on the American College Football network~\cite{Girvan2001CommunitySI}. Given embeddings $\{e_i\}_{i=1}^n$, we run $K$-means with $K{=}12$ (the number of ground-truth conferences) using Euclidean distance: we perform $K$-means on $\{e_i\}_{i=1}^n$ to obtain community assignments $c_i\in\{1,\ldots,K\}$ by minimizing
\begin{equation}
\min_{\{\mu_k\},\{c_i\}} \sum_{i=1}^{n}\left\|e_i-\mu_{c_i}\right\|_2^2,
\label{eq:ec_kmeans}
\end{equation}
where $\mu_k$ is the centroid of community $k$.
We report ARI and NMI against the ground-truth conference labels. Unless otherwise noted, we use $10$ random initializations and report the mean.

\paragraph{Influence maximization protocol.}
We instantiate influence maximization on the Berkeley road network under the Independent Cascade (IC) model with budget $k$. To define diffusion probabilities on a deterministic road graph, we assign each edge $(u,v)\in E$ an activation probability via a fixed monotone mapping of embedding similarity:
\[
p_{uv} \;=\; \mathrm{clip}\Big(\alpha\cdot \sigma(\mathrm{sim}(e_u,e_v))\,,\,0,\,1\Big),
\]
where $\mathrm{sim}(\cdot,\cdot)$ is cosine similarity, $\sigma(\cdot)$ is an optional squashing function (e.g., logistic), $\alpha$ is a global scaling parameter, and $\mathrm{clip}$ truncates to $[0,1]$. We keep the same mapping and parameters across all compared methods.  Finally, we evaluate expected spread under IC using Monte Carlo simulation with $R$ runs per seed set and report the mean spread.

\end{document}